\documentclass[10pt,twocolumn,letterpaper]{article}

\usepackage[pagenumbers]{cvpr} % To force page numbers, e.g. for an arXiv version
\usepackage{graphicx}
\usepackage{pifont}
\usepackage{amsmath}
\usepackage{enumitem}

\definecolor{cvprblue}{rgb}{0.21,0.49,0.74}
\usepackage[pagebackref,breaklinks,colorlinks,allcolors=cvprblue]{hyperref}
\usepackage{multirow}
\usepackage{verbatim}
\usepackage[accsupp]{axessibility}  % Improves PDF readability for those with disabilities.
\def\paperID{} % *** Enter the Paper ID here
\def\confName{CVPR}
\def\confYear{2026}

\title{SkyReels-Text: Fine-Grained Font-Controllable Text Editing for Poster Design}

\author{Yunjie Yu, Jingchen Wu, Junchen Zhu, Chunze Lin\footnotemark[1], Guibin Chen\\
Skywork AI\\
}

\begin{document}
\twocolumn[{
\renewcommand\twocolumn[1][]{#1}
\maketitle 
\vspace{-30pt}
\begin{center} 
\centering 
\includegraphics[width=1\textwidth]{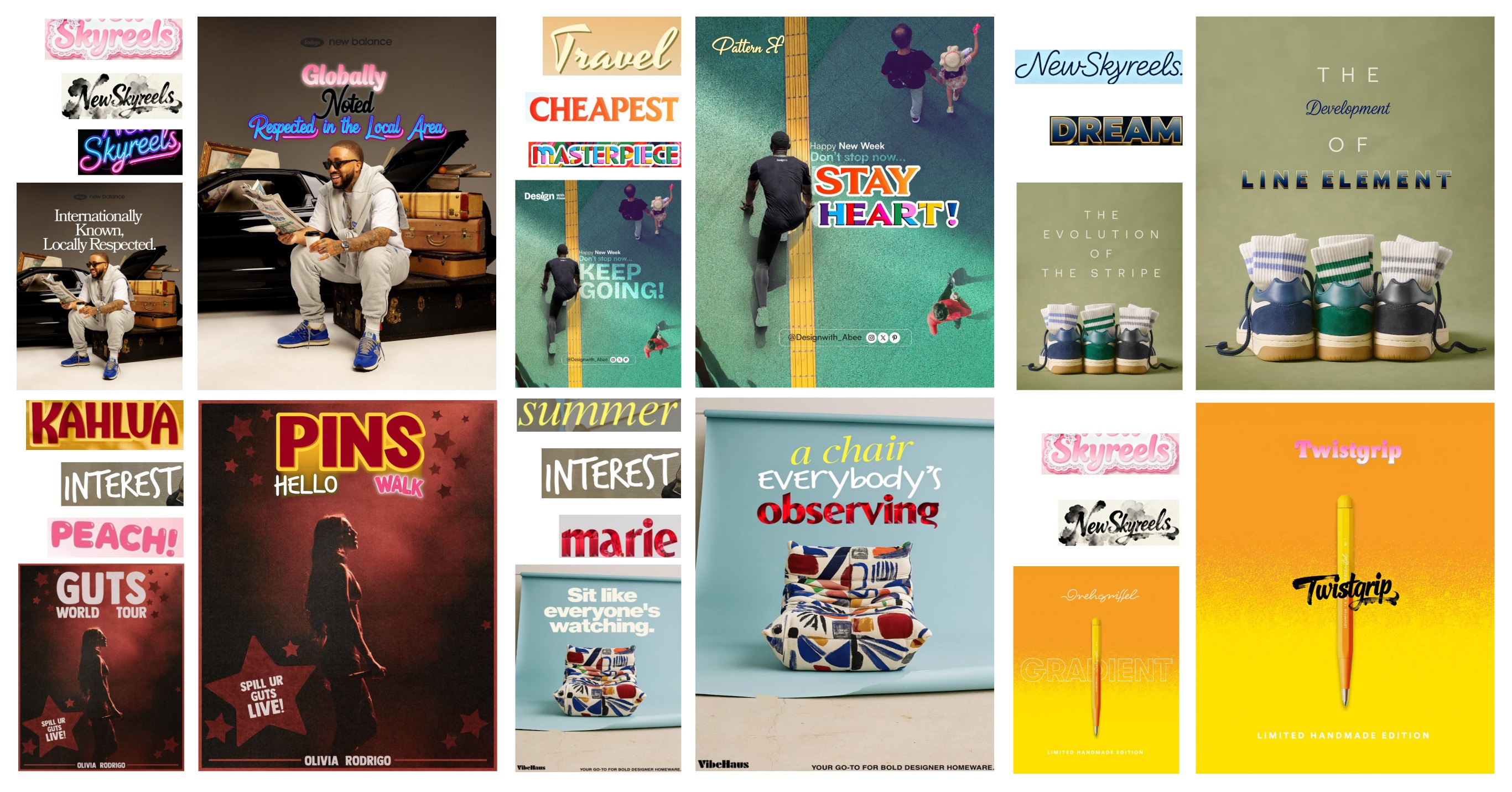} 
\captionof{figure}{\textbf{Font-controllable text editing.} SkyReels-Text modifies text within images using novel fonts guided by a single reference image.}
\label{fig:firstpage}
\end{center}
}]
\renewcommand{\thefootnote}{\fnsymbol{footnote}}
\footnotetext[1]{Corresponding author}
% \maketitle
\begin{abstract}
Artistic design, particularly poster design, often demands rapid yet precise modification of textual content while preserving visual harmony and typographic intent, especially across diverse font styles. Although modern image editing models have grown increasingly powerful, they still fall short in fine-grained, font-aware text manipulation, limiting their utility in professional workflows. To address this issue, we present SkyReels-Text, a novel font-controllable framework for precise poster text editing. Our method enables simultaneous editing of multiple text regions, each rendered in distinct typographic styles, while preserving the visual appearance of non-edited regions. Notably, our model requires neither font labels nor test-time fine-tuning: users can simply provide cropped glyph patches corresponding to their desired typography—even if the font is not included in any standard library. Extensive experiments on multiple benchmarks demonstrate that SkyReels-Text achieves state-of-the-art performance in both text fidelity and visual realism, offering unprecedented control over font families and stylistic nuances. This work bridges the gap between general-purpose image editing and professional-grade typographic design. Code and models are publicly available at \url{https://github.com/SkyworkAI/SkyReels-Text}.
\end{abstract}    
\section{Introduction}
\label{sec:intro}

Poster design is a cornerstone of visual communication in advertising, film, events, and social media. A key requirement in professional workflows is the ability to rapidly revise textual content—changing slogans, brand names, dates, or multilingual captions—while preserving the original visual harmony, typographic identity, and aesthetic intent. This demands not only accurate character rendering but also fine-grained control over font families and stylistic nuances, especially when handling complex scripts such as handwritten text.

Although modern diffusion-based image editing models (e.g., FLUX~\cite{flux2024, labs2025flux1kontextflowmatching}, Qwen-image~\cite{wu2025qwenimagetechnicalreport}, Seedream~4.0~\cite{seedream2025}) have achieved impressive results in general-purpose image manipulation, they remain fundamentally limited in their ability to perform fine-grained, font-aware text editing. These models can incorporate reference glyphs as visual context, but they often fail to correctly edit the textual content or accurately transplant the reference typography into the target region. As a result, either the semantic content is corrupted, or the generated text does not faithfully align with the desired position and layout—undermining the typographic precision essential for professional design. This limitation highlights a critical gap between current AI-based editing tools and the fine-grained control demanded by real-world design workflows.

Concurrently, another line of research has focused on developing specialized text editing approaches to address the challenge of accurate visual text generation. Models such as FLUX-Text~\cite{lan2025flux} and TextFLUX~\cite{xie2025textflux} inject explicit visual priors (e.g., rendered text content, positional masks) into powerful DiT backbones. Despite their effectiveness in spelling correctness, a critical drawback remains: they lack a mechanism to accept an arbitrary font style extracted from a user-provided visual patch. Consequently, they offer no fine-grained control over the output typography and cannot guarantee that the edited text will precisely mimic a given reference font—a paramount requirement in professional workflows such as poster design.

In this paper, we present SkyReels-Text, a novel framework for fine-grained, font-controllable text editing that enables precise textual content modification while preserving typographic style. Our method introduces a dual-stream visual conditioning mechanism that leverages user-provided glyph patches as explicit visual references, eliminating the need for font labels or test-time fine-tuning. By formulating text editing as a regional modification task with targeted visual conditioning, our approach provides unambiguous priors that ensure both semantic accuracy and typographic fidelity. This design enables the simultaneous editing of multiple text regions with heterogeneous typographic styles, supporting the complex multilingual layouts common in professional poster design. Unlike previous approaches that depend on text prompts alone or internal glyph priors, SkyReels-Text utilizes explicit visual cues to achieve unprecedented control over font families, weights, and stylistic nuances, effectively meeting the needs of professional-grade typographic design.

The key contributions of this work are fourfold:  
\begin{itemize}
\item A novel visual conditioning strategy that uses reference glyph patches to enable fine-grained, font-controllable text editing without requiring font labels.
\item A VLM-based OCR system capable of recognizing and localizing ornamental or calligraphic fonts, which are highly challenging for conventional OCR methods.
\item A dedicated pipeline integrating LLM text replacement, local editing, and dual-verification to ensure content-style decoupling, yielding a dataset of 100K aligned original–-edited image pairs.
\item A comprehensive evaluation on both public and in-house text editing benchmarks, achieving state-of-the-art performance in both semantic accuracy and typographic style fidelity.
\end{itemize}

\section{Related Work}
\subsection{Image Editing Models}
In recent years, with the advancement of image generation models, image editing has achieved increasingly sophisticated capabilities, including semantic manipulations such as object insertion, style transfer, and inpainting. FLUX.1 Kontext~\cite{labs2025flux1kontextflowmatching} extends the image generation model FLUX~\cite{flux2024} into a robust image editing framework by treating the input image as contextual information. Meanwhile, Qwen-Image-Edit~\cite{wu2025qwenimagetechnicalreport}, built on the Qwen-Image foundation model, specializes in precise image manipulation with native bilingual (Chinese and English) text rendering capabilities. Recently, multimodal foundation models have unified vision and language to enable open-domain, instruction-guided image editing. Models such as Gemini 2.5 Flash Image~\cite{gemini}, GPT-4o~\cite{gpt4o}, and Seedream~4.0~\cite{seedream2025} support image editing through natural language instructions and the fusion of multiple inputs, showcasing strong capabilities in multi-turn editing and large-scale conversational refinement. However, a key limitation of these general-purpose models is their primary focus on broad image editing tasks. Consequently, they lack the specific robustness required for reliable text editing, especially in font transfer scenarios. This fundamentally limits their capacity for the fine-grained, typographically precise editing demanded in professional artistic design workflows.

\subsection{Text Editing Models}
Specialized text synthesis methods have sought to bridge the gap between general image generation and typographic precision. Early works like Imagen~\cite{saharia2022photorealistic} and Ediff-i~\cite{balaji2022ediff} employed large-scale language models (e.g., T5-XXL~\cite{chung2024scaling}) to improve textual spelling accuracy. Subsequent approaches introduced glyph-aware conditioning: GlyphDraw~\cite{ma2023glyphdraw} and GlyphControl~\cite{yang2023glyphcontrol} used glyph images to guide character rendering, while TextDiffuser~\cite{chen2023textdiffuser} leveraged OCR-derived segmentation masks. Multilingual extensions such as AnyText~\cite{tuo2024anytext} and font-aware systems like DreamText~\cite{wang2025dreamtext} further enriched control over script and style. FLUX-Text~\cite{lan2025flux} and TextFLUX~\cite{xie2025textflux}, built on the FLUX foundation model, inject glyph priors (rendered glyphs and position masks) into the DiT backbone to enhance its text generation capability. In contrast, our work departs from this paradigm: rather than depending on text prompts alone or internal glyph priors, SkyReels-Text leverages user-provided glyph patches as visual references to achieve font-controllable text editing.

\section{Method}
\label{sec:method}

\subsection{Preliminaries}
Given a source poster image $X_{\text{src}}$, a set of text instances $T = \{(t_1, r_1, g_1), (t_2, r_2, g_2), \dots, (t_n, r_n, g_n)\}$, and a text prompt $y$ containing the target text provided by the user, our goal is to generate an edited image $X_{\text{out}}$ where the original text regions are replaced with the target text in visually consistent fonts. Here, $t_i$ represents the $i$-th target text to be edited within region $r_i$ using the reference glyph $g_i$, and $n$ denotes the total number of text instances to be edited.

\subsection{Data Processing}
\textbf{Data Curation.} Most text editing methods rely on synthetic data generated by rendering text with standard system fonts (e.g., Arial, SimSun) onto simple backgrounds. While scalable, this approach suffers from three critical limitations: (1) limited font diversity, excluding rich real-world styles such as calligraphic or custom designs; (2) unrealistic backgrounds, lacking the texture, lighting, and layout complexity found in professional designs; and (3) unnatural glyph-background integration, creating a significant domain gap with real-world editing scenarios.

To address these issues, we curate a high-quality dataset from real-world posters collected from design platforms and public repositories. We apply a multi-stage data curation pipeline: (i) filtering out low-resolution posters; (ii) discarding images without legible text; and (iii) employing a custom-trained vision-language model (VLM) as a robust OCR system to accurately detect and recognize text—particularly ornamental or calligraphic fonts that are highly challenging for conventional OCR methods. This real-world, data-centric approach ensures that our model learns authentic font-background interactions and generalizes well to professional design contexts.

\noindent\textbf{VLM-Based OCR Model.} Existing OCR systems, ranging from practical models such as PP-OCRv4~\cite{ppocrv4} and PP-OCRv5~\cite{cui2025paddleocr30technicalreport} to the more powerful VLM-based approach PaddleOCR-VL~\cite{cui2025paddleocrvlboostingmultilingualdocument}, are typically optimized for clean, regular typefaces and scene text. Thus, they fail to generalize to ornate calligraphic scripts or custom-designed letterforms that exhibit irregular strokes and non-standard spatial layouts. Given the complexity of recognizing non-standard fonts, which may be classified as icons or patterns, we introduce a specialized OCR model to enable simultaneous analysis from both visual and semantic perspectives. Specifically, we fine-tune the Qwen2.5-VL 7B~\cite{bai2025qwen25vltechnicalreport} model, which excels at parsing non-standard textual patterns within images. This adaptation allows the model to better handle diverse font geometries and artistic variations.

\begin{figure}[t]
    \centering
    \includegraphics[width=\linewidth]{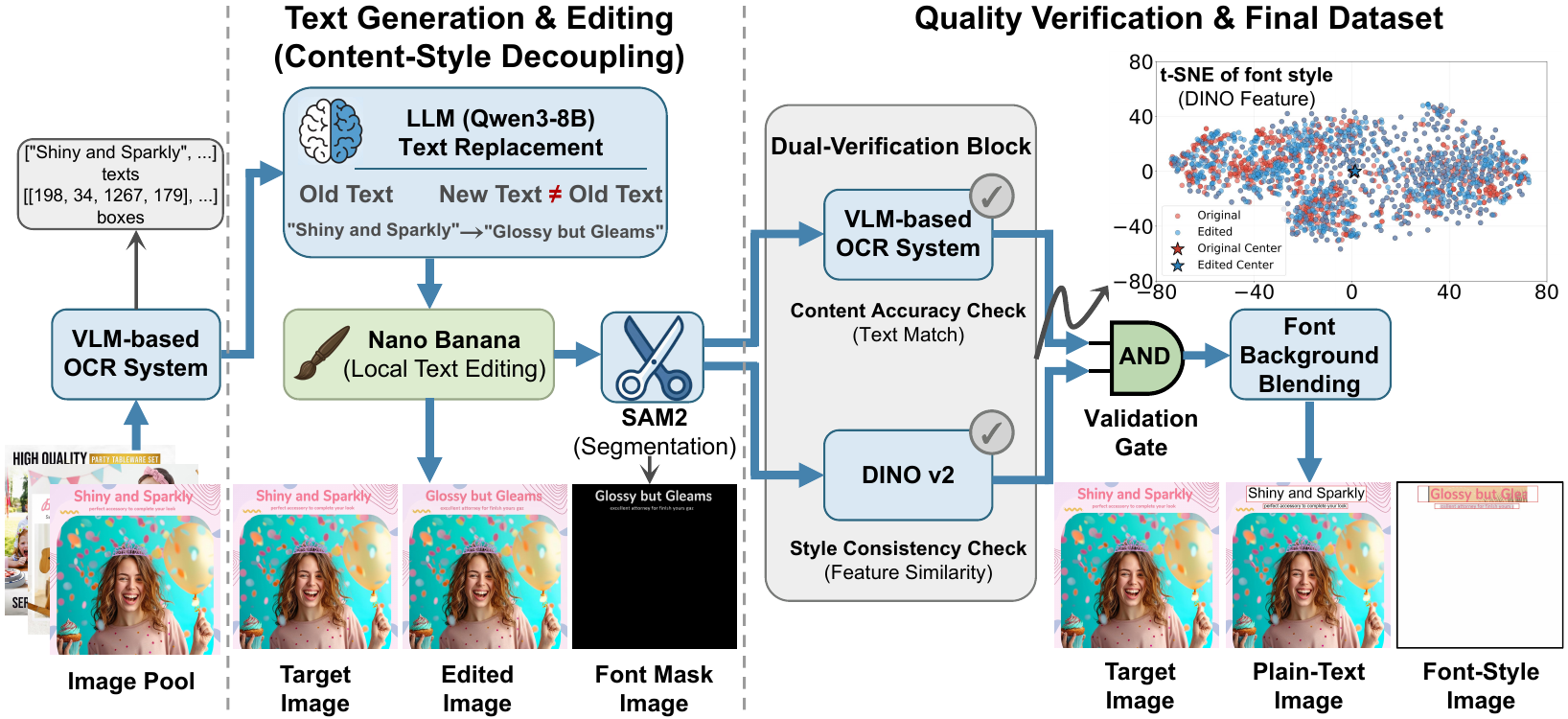}
    \caption{\textbf{Pipeline of content-style decoupling.} We utilize our VLM-based OCR and Qwen3-8B to obtain divergent text content, followed by local editing via Nano Banana and mask extraction via SAM2 to generate decoupled image pairs. A rigorous Dual-Verification Block ensures both content accuracy and style consistency. The t-SNE visualization of font styles verifies the style alignment as well as the high diversity of our dataset.}
    \label{fig:data_generation}
\end{figure}

\begin{figure*}[t]
    \centering
    \includegraphics[width=\linewidth]{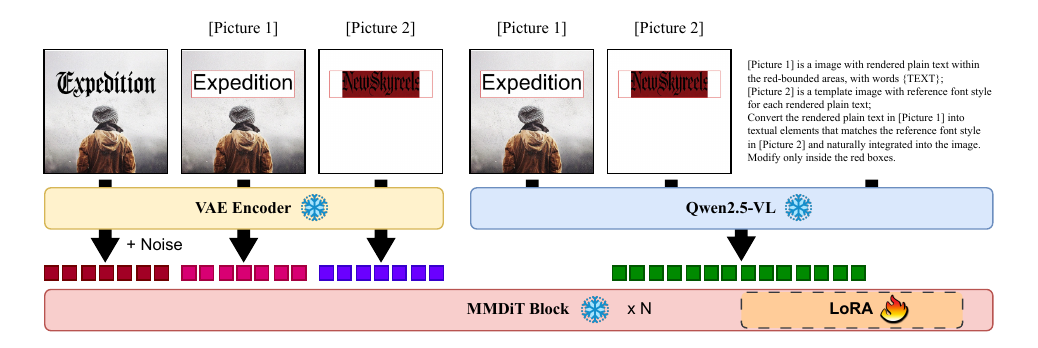}
    \caption{\textbf{Overview of the proposed method.} SkyReels-Text employs a dual-stream visual conditioning mechanism to decouple textual content and typographic style. Specifically, the VAE encoder extracts visual features from the plain-text reference and the glyph map, which are then concatenated with the noisy image latent. Simultaneously, Qwen2.5-VL processes the multimodal editing instructions. The combined sequence is fed into MMDiT blocks equipped with trainable LoRA modules, enabling zero-shot style transfer.}
    \label{fig:overview}
\end{figure*}

\noindent\textbf{Data Generation with Content-Style Decoupling.} A critical challenge in style transfer learning is content interference. Specifically, the model often conflates content and style features for certain characters within the text. To address this issue, we designed a dedicated data generation pipeline, as illustrated in Figure~\ref{fig:data_generation}. First, we identify high-quality text instances in rich-text posters using our VLM-based OCR model. During this step, we filter out overlapping text regions. For the remaining text, we further employ Qwen3-8B~\cite{yang2025qwen3} to generate semantically divergent replacement text. Crucially, we ensure the new word set is disjoint from the original text to maximize content decoupling. Next, we leverage the Nano Banana~\cite{gemini} editing model to replace the original content while preserving all typographical attributes, including font, color, and local alignment, thereby ensuring the retention of fine-grained style patterns. To rigorously validate the fidelity of this process, we implement a two-pronged verification strategy. First, we re-apply our VLM-based OCR model to confirm the content accuracy of the newly rendered text. Second, we extract DINO v2 features~\cite{oquab2023dinov2} from both the original and edited text regions to verify high feature-level similarity, confirming style preservation. This methodology produces perfectly aligned original--edited image pairs that share identical styles but contain maximally different content. Finally, this robust pipeline yields a high-quality dataset of 100K samples, comprising paired target images and reference fonts where the text content and style are completely decoupled.

\subsection{Font-Controllable Text Editing}
General image editing models typically rely on textual prompts to modify text within an image. However, such prompt-driven approaches suffer from inherent limitations in fine-grained precision, often resulting in failed modifications, particularly the inaccurate reproduction of original typographic styles. To address this, we formulate font-controllable text editing as a regional modification task, where the model is conditioned to modify a user-specified region $r_i$ based on a target text $t_i$ and a reference glyph $g_i$.

To overcome the limitations of purely text-based guidance, we propose a diffusion model with a dual-stream visual conditioning mechanism that provides explicit, unambiguous priors for precise typographic control. As demonstrated in Figure~\ref{fig:overview}, we decouple content and style into two distinct conditions. First, we generate a plain-text reference $X_{\text{text}}$ (Picture 1) by inpainting the masked text regions in the original input with the target text $t_i$, thereby establishing the desired textual content and layout within the visual context. Second, we create a dedicated glyph map $X_{\text{glyph}}$ (Picture 2) by rendering the corresponding glyph $g_i$ onto a canvas matching the input size. By conditioning the model on both $X_{\text{text}}$ and $X_{\text{glyph}}$, we ensure that the generated output achieves precise alignment with the intended semantic content and the specified typographic style.

Instead of injecting glyph priors via parameter-heavy embedding modules or ControlNet, we directly use a frozen VAE encoder to extract features from the plain-text reference $X_{\text{text}}$ and the glyph map $X_{\text{glyph}}$. This simplifies the training process while ensuring effective extraction and compression of visual information. These VAE features are subsequently injected into the model by concatenating them with the noisy image latent $z_t$ along the sequence dimension:
\[
Z_{\text{in}} = \text{Concat}(z_t, \text{VAE}(X_{\text{text}}), \text{VAE}(X_{\text{glyph}})).
\]
This design allows the model to directly attend to text content and font exemplars, enabling zero-shot style transfer.

\subsection{Text-Region Weighted Loss}
Generally, text regions occupy only a small fraction of the overall image, typically less than 10\%, which means that applying uniform optimization across all pixels can hinder the convergence of text generation during training. To ensure high fidelity in the edited text regions while preserving background integrity, we introduce a text-aware weighted reconstruction loss. 
Let $X_{\text{gt}}$ be the ground-truth edited image and $\hat{X}$ be the model prediction. We compute the standard diffusion loss (e.g., $\mathcal{L}_2$ or flow matching loss) but reweight it spatially using the text mask $M$:

\[
\mathcal{L} = \mathbb{E}\left[ \| X_{\text{gt}} - \hat{X} \|^2_2 \odot \left(1 + \lambda \cdot M \right) \right],
\]
where $\lambda > 0$ is a hyperparameter that amplifies the loss in text regions, and $M$ is a tight text-region mask generated by SAM2~\cite{ravi2024sam}. This simple yet effective strategy focuses optimization on glyph accuracy without requiring a separate perceptual encoder or multi-stage training. In practice, we set $\lambda = 5$, which yields an optimal balance between text fidelity and background consistency.

\subsection{Distillation for Fast Inference}
To accelerate inference without sacrificing font-controllable editing quality, we distill SkyReels-Text using the Improved Distribution Matching Distillation (DMD2) paradigm~\cite{yin2024improved}. We adapt DMD2 to our architecture, producing an 8-step student model that closely matches the teacher's output distribution while preserving fine-grained font awareness. This reduces the number of sampling steps and eliminates the need for classifier-free guidance, achieving a speedup of over $10\times$ during inference.

\section{Experiments}
\subsection{Implementation Details}
\label{sec:implementation_details}
We adopt the Qwen-Image-Edit~\cite{wu2025qwenimagetechnicalreport} model as our baseline to ensure high-fidelity and reliable editing performance. To preserve the strong capabilities of the underlying foundation model, we fine-tune it using LoRA with a rank of 64. We set the batch size to 64, employ the AdamW optimizer with a learning rate of $10^{-4}$, and train for 2 epochs, consuming about 512 A100-hours. For the VLM-based OCR model, we fine-tune Qwen2.5-VL 7B~\cite{bai2025qwen25vltechnicalreport} with a learning rate of $10^{-4}$ using approximately 72 A100-hours, enabling precise and pixel-localized detection of textual content.

\subsection{Evaluation Dataset}
We evaluate the performance of SkyReels-Text across artistic, general, and handwritten text synthesis using one in-house benchmark and two categories of public benchmarks. Specifically, we meticulously curate a benchmark of 200 diverse text samples encompassing a wide range of typographic styles, referred to as the SkyReels-Text Benchmark. For assessing the accuracy and visual fidelity of synthesized text, we utilize the AnyText benchmark~\cite{tuo2024anytext}, which comprises 1,000 images from the Wukong~\cite{gu2022wukong} (Chinese) and LAION~\cite{schuhmann2021laion} (English) datasets. To validate the model's generalization capabilities for handwritten text generation, we employ two widely used datasets: IAM~\cite{marti2002iam} and CVL~\cite{kleber2013cvl}. For the IAM dataset, we select text lines from 161 writers following the protocol established by CSA-GAN~\cite{kang2021content}. For the CVL dataset, we construct the test set from 27 writers following the standard CVL split.

\begin{table}[t]
    \centering
    \caption{Performance comparison on SkyReels-Text benchmark.}
    \label{tab:SkyReels_text_performance}
    \resizebox{\linewidth}{!}{
        \begin{tabular}{l|ccccc} 
            \toprule
             Method & Sen.~Acc$\uparrow$ & NED$\uparrow$ & Spatial$\uparrow$ & DINO$\uparrow$ & B-PSNR$\uparrow$ \\
            \midrule % 保留表头下的横线
            Nano Banana~\cite{gemini}  & 0.7290 & 0.9195 & 0.7011 & 0.8125 & 28.78 \\
            Seedream~4.0~\cite{seedream2025}  & 0.7772 & 0.9348 & 0.6844 & 0.7629 & 25.50 \\
            FLUX-Kontext-Pro~\cite{labs2025flux1kontextflowmatching}  & 0.6390 & 0.8458 & 0.5063 & 0.8130 & 27.21 \\
            Qwen-Image-Edit~\cite{wu2025qwenimagetechnicalreport} & 0.7760 & 0.9337 & 0.5845 & 0.8209 & 25.88 \\
            FLUX-Text~\cite{lan2025flux} & 0.8266 & 0.9352 & 0.7503 & 0.6679 & \textbf{34.53} \\
            Calligrapher~\cite{ma2025calligrapherfreestyletextimage} & 0.6404 & 0.8811 & 0.7281 & 0.7162 & 24.26 \\
            \midrule
            \textbf{SkyReels-Text} & \textbf{0.8334} & \textbf{0.9502} & \textbf{0.7506} & \textbf{0.8503} & 34.17 \\
            \bottomrule
        \end{tabular}
    }
\end{table}

\begin{figure*}[t]
    \centering
    \includegraphics[width=\linewidth]{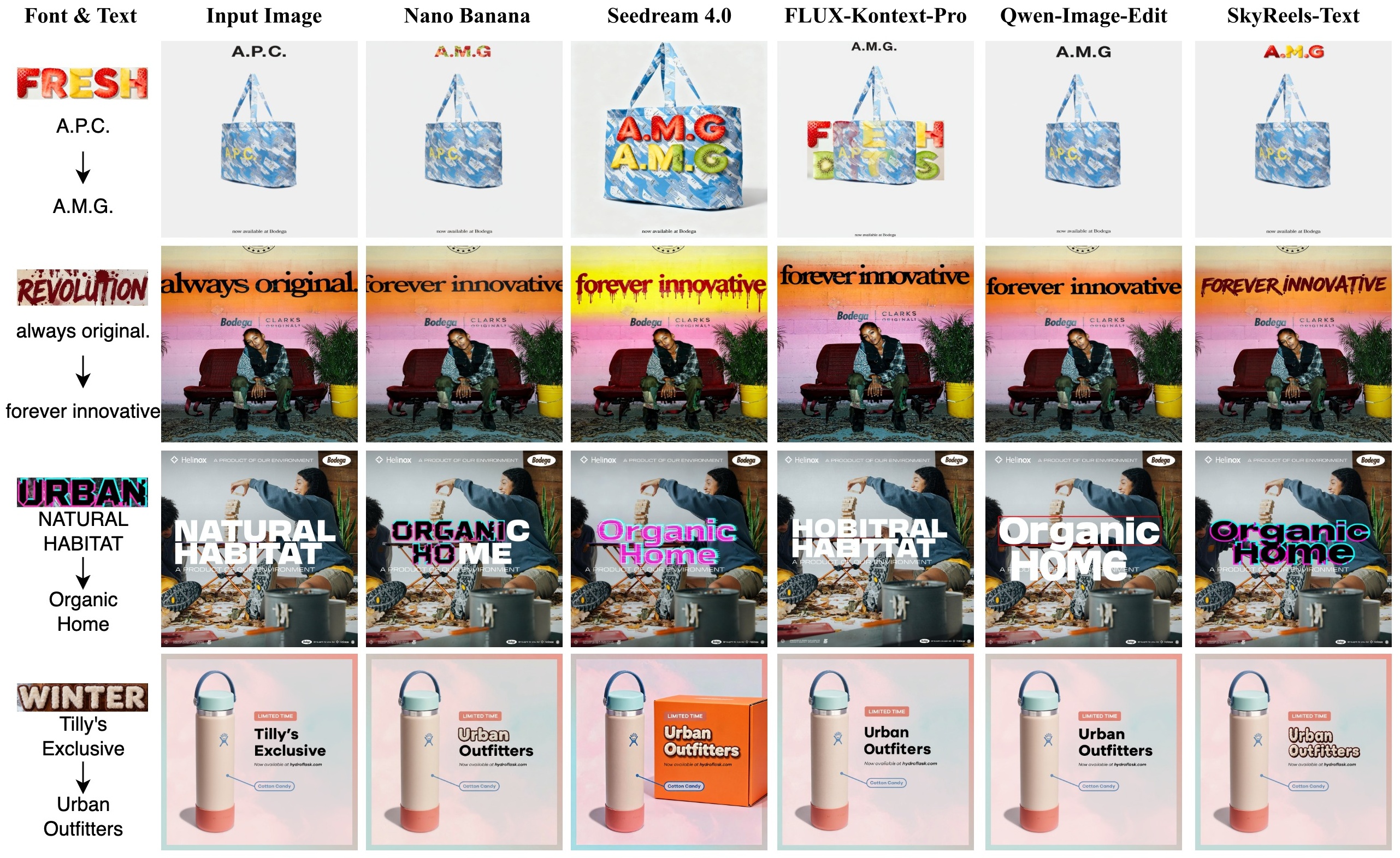}
    \caption{\textbf{Comparison with SOTA commercial image editing models in single-font editing.} The leftmost column presents the reference font image and the source/target plain text. SkyReels-Text produces edits that faithfully follow the target typography while preserving the background structure and content intact.}
    \label{fig:singlecomp}
\end{figure*}

\begin{table*}[t]
\centering
\caption{Performance comparison of different methods on AnyText benchmark for English and Chinese datasets.}
\label{tab:performance}
\begin{tabular}{l|cccc|cccc}
\toprule
\multirow{2}{*}{{Methods}} & \multicolumn{4}{c|}{{English}} & \multicolumn{4}{c}{{Chinese}} \\
% \cline{3-10}
 & {Sen.~Acc$\uparrow$} & {NED$\uparrow$} & {FID$\downarrow$} & {LPIPS$\downarrow$} & {Sen.~Acc$\uparrow$} & {NED$\uparrow$} & {FID$\downarrow$} & {LPIPS$\downarrow$} \\
\midrule
DiffSTE~\cite{ji2023improving}            & 0.4523 & 0.7814 & 52.74  & 0.1816 & 0.0363 & 0.1226 & 57.49  & 0.1276 \\
TextDiffuser~\cite{chen2023textdiffuser}        & 0.5176 & 0.7618 & 29.76  & 0.1564 & 0.0559 & 0.1218 & 34.19  & 0.1252 \\
DiffUTE~\cite{chen2023diffute}             & 0.4054 & 0.7005 & 25.35  & 0.1640 & 0.2978 & 0.5744 & 29.08  & 0.1745 \\
Anytext~\cite{tuo2024anytext}            & 0.6843 & 0.8588 & 21.59  & 0.1106 & 0.6476 & 0.8210 & 20.01  & 0.0943 \\
TextCtrl~\cite{zeng2024textctrl}           & 0.5853 & 0.8146 & 35.73  & 0.1978 & 0.3580 & 0.6084 & 49.79  & 0.2298 \\
Anytext2~\cite{tuo2024anytext2}      & 0.7915 & 0.9100 & 29.76  & 0.1734 & 0.7022 & 0.8420 & 26.52  & 0.1444 \\
FLUX-Text~\cite{lan2025flux}             & {0.8419} & {0.9400} & {13.85} & {0.0729} & {0.7132} & {0.8510} & {13.68} & {0.0541} \\
Calligrapher~\cite{ma2025calligrapherfreestyletextimage} & {0.7761} & {0.9011} & {22.41} & {0.0851} & {0.4369} & {0.6884} & {24.93} & {0.0797} \\
\midrule
Baseline                  & 0.8292 & 0.8708 & 35.62  & 0.2499 & 0.6735 & 0.8323 & 19.26  & 0.1239 \\
\textbf{SkyReels-Text}  & \textbf{0.8536} & \textbf{0.9406} & \textbf{6.12} & \textbf{0.0246} & \textbf{0.7710} & \textbf{0.8764} & \textbf{5.44} & \textbf{0.0192} \\
\bottomrule
\end{tabular}
\end{table*}

\begin{figure*}[t]
    \centering
    \includegraphics[width=0.95\linewidth]{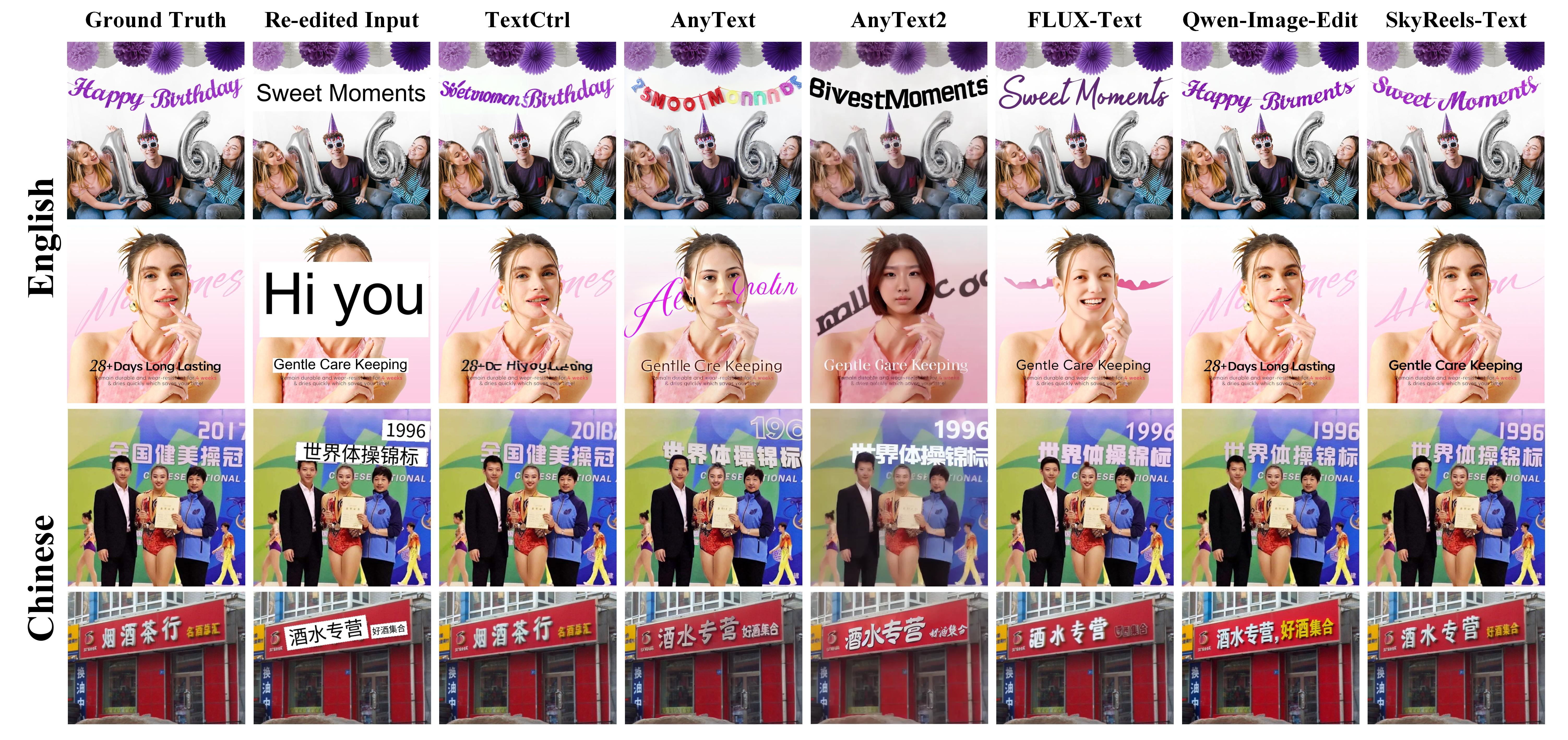}
    \caption{\textbf{Comparison with SOTA open-source models.} Visual results of scene text editing for both English and Chinese.}
    \label{fig:multicomp_opensource}
\end{figure*}

\subsection{Evaluation Metrics}
\label{sec:metrics}
For the SkyReels-Text benchmark, we evaluate performance across four key dimensions: (1) Text fidelity: Sentence Accuracy (Sen.~Acc.) and Normalized Edit Distance (NED = $1 - \text{Edit Distance}$), computed using our fine-tuned Qwen2.5-VL 7B model; (2) Layout alignment: spatial IoU between OCR-detected and ground-truth text boxes; (3) Font style adherence: DINO similarity between generated and reference text regions; and (4) Background preservation: B-PSNR, computed on non-text regions to quantify structural fidelity. For the AnyText benchmark~\cite{tuo2024anytext}, we adopt Sen.~Acc. and NED (evaluated via DuGuangOCR~\cite{duguangocr2023} for fair comparison) for text accuracy, along with Fréchet Inception Distance (FID)~\cite{Seitzer2020FID} and Learned Perceptual Image Patch Similarity (LPIPS)~\cite{zhang2018unreasonable} for distributional and perceptual image quality. For handwritten text generation, we use Handwriting Distance (HWD)~\cite{pippi2023hwd} to assess style fidelity, alongside Inception Score (IS)~\cite{salimans2016improved} and Geometry Score (GS)~\cite{khrulkov2018geometry} for overall visual quality.

\begin{table}[t]
\centering
\caption{Performance comparison on IAM and CVL datasets. For simplicity, we omit $\times 10^{-3}$ in all GS metrics.}
\label{tab:reduced_performance}
\resizebox{\linewidth}{!}{
\begin{tabular}{llcccc}
\toprule
Datasets & {Method} & Shot & HWD$\downarrow$ & IS$\uparrow$ & GS$\downarrow$ \\
\hline
\multirow{7}{*}{IAM} 
& TS-GAN~\cite{davis2020text}       & one & 2.11 & 1.76 & 28.7 \\
& CSA-GAN~\cite{kang2021content}     & few & 2.25 & 1.74 & 16.2 \\
& VATr~\cite{pippi2023handwritten}        & few & 1.87 & 1.69 & 14.5 \\
& DiffusionPen~\cite{nikolaidou2024diffusionpen} & few & 1.72 & 1.83 & 6.42 \\
& One-DM~\cite{dai2024one}       & one & 1.80 & 1.82 & 8.42 \\
& DiffBrush~\cite{dai2025beyond}    & one & {1.41} & {1.85} & 2.35 \\
& \textbf{SkyReels-Text} & one & \textbf{1.32} & \textbf{1.90} & \textbf{1.26} \\
\hline
\multirow{6}{*}{CVL} 
& CSA-GAN~\cite{kang2021content}     & few & 1.72 & 1.48 & 671 \\
& VAtR~\cite{pippi2023handwritten}        & few & 1.50 & 1.44 & 143 \\
& DiffusionPen~\cite{nikolaidou2024diffusionpen} & few & 1.32 & 1.59 & 50.8 \\
& One-DM~\cite{dai2024one}       & one & 1.47 & 1.46 & 129 \\
& DiffBrush~\cite{dai2025beyond}    & one & {1.06} & {1.70} & \textbf{29.6} \\
& \textbf{SkyReels-Text} & one & \textbf{0.89} & \textbf{1.71} & 31.0 \\
\bottomrule
\end{tabular}
}
\end{table}

\subsection{Comparison with State-of-the-Art Methods}
\textbf{Quantitative Results.} We evaluate SkyReels-Text on our comprehensive benchmark against the current state-of-the-art (SOTA) commercial and open-source methods, including Nano Banana~\cite{gemini}, Seedream~4.0~\cite{seedream2025}, FLUX-Kontext-Pro~\cite{labs2025flux1kontextflowmatching}, Qwen-Image-Edit~\cite{wu2025qwenimagetechnicalreport}, FLUX-Text~\cite{lan2025flux}, and Calligrapher~\cite{ma2025calligrapherfreestyletextimage}. As shown in Table~\ref{tab:SkyReels_text_performance}, SkyReels-Text achieves high text fidelity with a Sen.~Acc. of 83.34\% and an NED of 95.02\%. Crucially, it demonstrates high layout alignment and font style adherence, as evidenced by the spatial IoU and DINO scores, respectively. Furthermore, the model's ability to maintain background integrity, quantified by B-PSNR, is comparable to that of FLUX-Text~\cite{lan2025flux}. Overall, compared to these competing methods, SkyReels-Text achieves SOTA performance, highlighting its superiority in text quality and font consistency.

Next, we evaluate our model on the AnyText benchmark and compare it against the current SOTA methods, including DiffSTE~\cite{ji2023improving}, TextDiffuser~\cite{chen2023textdiffuser}, DiffUTE~\cite{chen2023diffute}, AnyText~\cite{tuo2024anytext}, TextCtrl~\cite{zeng2024textctrl}, AnyText2~\cite{tuo2024anytext2}, FLUX-Text~\cite{lan2025flux}, and Calligrapher~\cite{ma2025calligrapherfreestyletextimage}. As shown in Table~\ref{tab:performance}, SkyReels-Text achieves the best performance across all metrics (Sen.~Acc., NED, FID, and LPIPS) on both the Chinese and English datasets. Specifically, SkyReels-Text achieves an English Sen.~Acc. of 85.36\% and a Chinese Sen.~Acc. of 77.10\%, surpassing FLUX-Text by 1.37\% and 7.50\%, respectively. Regarding the NED metric, it also attains the highest scores of 94.06\% for English and 87.64\% for Chinese. Furthermore, SkyReels-Text yields the lowest FID and LPIPS scores, indicating superior distribution-level and perceptual alignment with real images.

\begin{figure*}[t]
    \centering
    \includegraphics[width=0.9\linewidth]{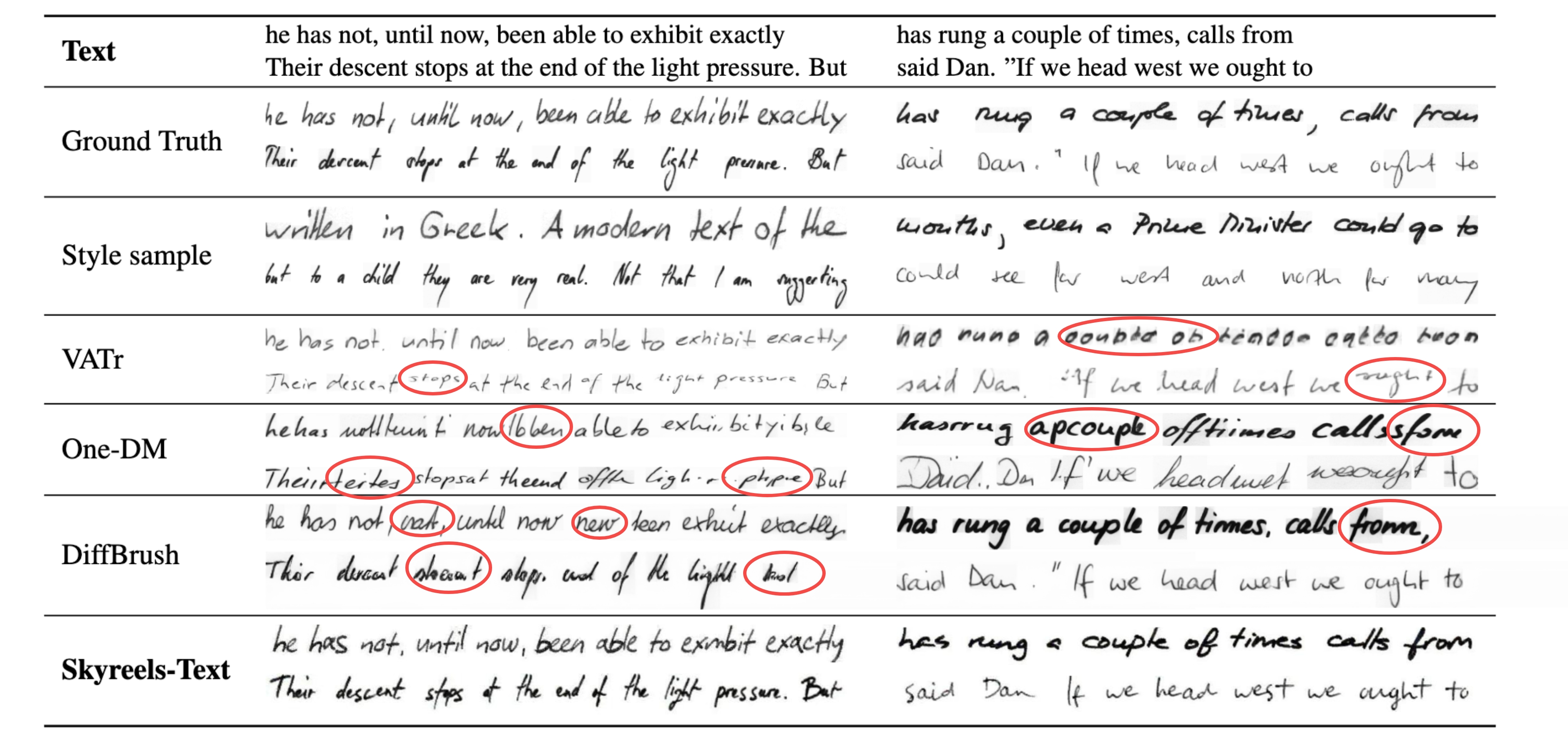}
    \caption{\textbf{Handwritten text generation.} Qualitative comparison with SOTA approaches using challenging exemplars.}
    \label{fig:comparison handwrite}
\end{figure*}

Finally, we evaluate our model on handwritten text datasets. Although we do not explicitly train on dedicated handwritten data, our model can still generate text images that faithfully mimic the target handwritten style when provided with reference images. As shown in Table~\ref{tab:reduced_performance}, we compare our model against the latest handwritten text generation models. SkyReels-Text achieves highly competitive results compared to models specifically designed and trained on handwritten data. In particular, compared to DiffBrush~\cite{dai2025beyond}, our method attains lower HWD scores of 1.32 and 0.89 on the IAM and CVL datasets, respectively—indicating superior style imitation capabilities. This demonstrates that SkyReels-Text can generate highly realistic handwritten text line images without any additional training, relying solely on the provided reference samples.

\noindent \textbf{Qualitative Results.}
To further evaluate our model, we present qualitative comparisons against both commercial and open-source models. First, we select four SOTA commercial image editing models: Nano Banana, Seedream~4.0, FLUX-Kontext-Pro, and Qwen-Image-Edit. To demonstrate the superiority of SkyReels-Text, we evaluate both single-font and multi-font editing, as shown in Figure~\ref{fig:singlecomp} and the supplement. SkyReels-Text successfully transfers reference font styles and performs accurate text edits in both settings, whereas the competing models either fail to apply the intended edits or struggle to preserve the background content. Second, we compare our model against open-source models. As shown in Figure~\ref{fig:multicomp_opensource}, SkyReels-Text preserves the original font style while accurately modifying the textual content, achieving precise typographic consistency for both English and Chinese. Evaluations on samples beyond the poster domain demonstrate our model's robustness to diverse real-world images.

In addition to evaluating the text editing, we present a detailed comparison focusing on pure font editing. Specifically, we select challenging handwritten fonts as exemplars. As shown in Figure~\ref{fig:comparison handwrite}, although our model is not explicitly trained on handwritten datasets, it successfully generalizes to handwritten text using only a single reference. Even compared to methods specifically designed and trained for handwritten text generation, SkyReels-Text achieves superior preservation of the font style and generates text that more closely aligns with the ground truth. We provide additional comparisons on Chinese handwritten text in the supplement. This strong generalization and zero-shot transfer capability obviates the need for extensive domain-specific fine-tuning, offering a flexible, efficient, and scalable solution for generating stylized text images.

\begin{figure}[t]
    \centering
    \includegraphics[width=\linewidth]{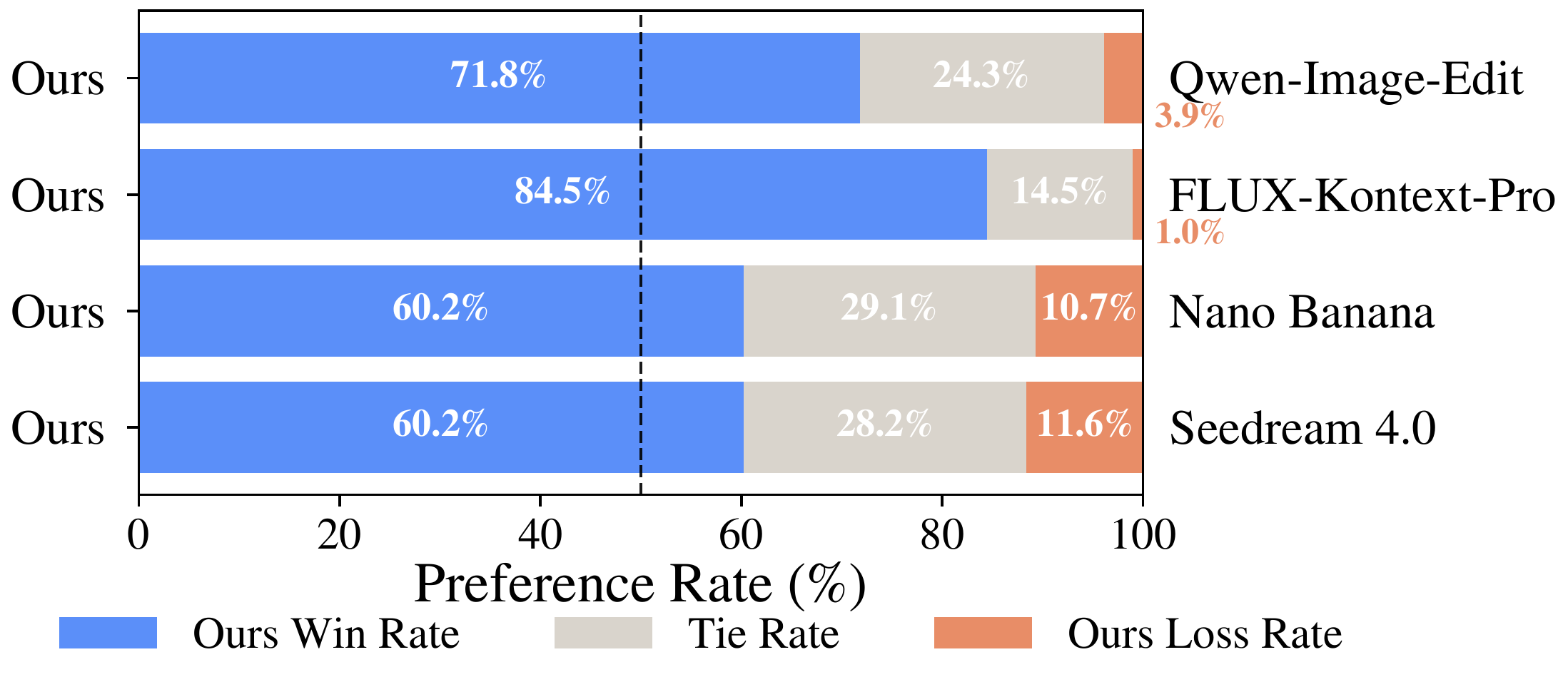}
    \caption{\textbf{User preference study.} SkyReels-Text outperforms all competitors in human preference.}
    \label{fig:user_study}
\end{figure}

\noindent \textbf{User Study.} To further validate the effectiveness of SkyReels-Text, we conducted a user study comparing our model with four commercial image editing systems: Qwen-Image-Edit, FLUX-Kontext-Pro, Nano Banana, and Seedream~4.0. Figure~\ref{fig:user_study} indicates that SkyReels-Text consistently outperforms all competitors. In particular, it achieves clear advantages over Qwen-Image-Edit and FLUX-Kontext-Pro, with win rates of 71.8\% and 84.5\%, respectively. When compared to Nano Banana and Seedream~4.0, SkyReels-Text still maintains a substantial advantage, achieving a 60.2\% win rate against a mere $\sim$10\% loss rate. These results underscore that SkyReels-Text provides a more reliable and user-favored editing experience than leading commercial alternatives.

\subsection{Ablation Studies}
We conduct ablations on the SkyReels-Text benchmark to evaluate our key components: the font style reference and the text-region weighted loss.

\begin{table}[t]
    \centering
    \caption{Ablation experiments for Font Style Reference (FSR) and weight $\lambda$ of Text-Region Weighted Loss. We use \textbf{boldface} and \underline{underlining} to denote the best and second-best values.}
    \label{tab:ablation_fsr_lambda}
    \resizebox{\linewidth}{!}{
        \begin{tabular}{c|ccccc} 
            \toprule
             FSR /$\lambda$  &  Sen.~Acc$\uparrow$ & NED$\uparrow$ & Spatial$\uparrow$ & DINO$\uparrow$ & B-PSNR$\uparrow$ \\
            \midrule % 保留表头下的横线
            \texttimes & \textbf{0.9327} & \textbf{0.9734} & 0.6986 & 0.6995 & 33.61 \\
            \checkmark & 0.8334 & 0.9502 & \textbf{0.7506} & \textbf{0.8503} & \textbf{34.17} \\
            \midrule
            $\lambda = 0$ & 0.7998 & 0.9378 & 0.7501 & 0.8473 & \textbf{34.28} \\
            $\lambda = 1$ & 0.8126 & 0.9392 & 0.7492 & 0.8484 & \underline{34.25} \\
            $\lambda = 2$ & 0.8267 & 0.9426 & \underline{0.7505} & 0.8496 & 34.21 \\
            $\lambda = 5$ & \underline{0.8334} & \underline{0.9502} & \textbf{0.7506} & \textbf{0.8503} & 34.17 \\
            $\lambda = 10$ &  \textbf{0.8363} & \textbf{0.9558} & 0.7503 & \underline{0.8502} & 33.13 \\
            \bottomrule
        \end{tabular}
    }
\end{table}

\noindent \textbf{Font Style Reference.} As shown in Table~\ref{tab:ablation_fsr_lambda}, incorporating the Font Style Reference (FSR) substantially enhances the model's ability to align with the target style, as evidenced by a higher DINO score. In contrast, the setting without FSR yields better Sen.~Acc and NED for two reasons: first, the task becomes simpler under single-condition generation; second, current OCR systems struggle to parse highly stylized text when FSR is enabled. Furthermore, the improved spatial score with FSR is driven by the fact that the reference image—cropped from the original—provides implicit spatial cues that facilitate accurate placement of generated text relative to the ground truth layout.

\noindent \textbf{Text-Region Weighted Loss.} Table~\ref{tab:ablation_fsr_lambda} also evaluates the impact of the weight $\lambda$ in the text-region weighted loss. Since text occupies only a small portion of the image, setting $\lambda > 0$ is essential to direct optimization toward glyph fidelity. Increasing $\lambda$ leads to consistent improvements in Sen.~Acc and NED, confirming the benefit of emphasizing text regions. While $\lambda = 10$ achieves the highest text accuracy, $\lambda = 5$ provides the optimal balance, maximizing spatial IoU and DINO scores—critical indicators of layout and style adherence. Thus, we set $\lambda = 5$ for the best trade-off between text convergence and background fidelity.

\section{Conclusion}
In this paper, we introduce SkyReels-Text, a novel framework for fine-grained, font-controllable text editing. It enables users to simultaneously modify multiple text regions, each guided by a user-provided reference glyph. We demonstrated that SkyReels-Text achieves state-of-the-art zero-shot performance while preserving background integrity, entirely bypassing the need for font labels or additional training. This work not only provides a highly practical solution for real-world design workflows but also paves the way for fine-grained, reference-driven visual generation.
{
    \small
    \bibliographystyle{ieeenat_fullname}
    \bibliography{main}
}
\clearpage
\maketitlesupplementary

\appendix % <--- 关键：将后续的 \section 编号切换为 A, B, C...

This supplementary material provides further details on the SkyReels-OCR dataset (Section~\ref{sec:ocr_dataset}) and the content-style decoupled dataset (Section~\ref{sec:content_style_decoupled_dataset}), and presents more qualitative results (Section~\ref{sec:more_qualitative_results}) to complement the main manuscript.

\section{SkyReels-OCR Dataset}
\label{sec:ocr_dataset}
Existing OCR systems are optimized for regular scene text but struggle with ornate, non-standard font geometries. Therefore, we adopt a VLM-based OCR model, Qwen2.5-VL 7B~\cite{bai2025qwen25vltechnicalreport}, for improved parsing of complex textual patterns. To address the lack of training data across diverse font geometries and artistic variations, we propose the SkyReels-OCR dataset, a meticulously curated collection of high-quality real data sourced from professional design platforms (i.e., publicly accessible images from Pinterest and Amazon, collected solely for academic research) and a public repository, namely the AnyWord-3M~\cite{tuo2024anytext} dataset. Our dataset covers a diverse range of text styles, including regular, handwritten, calligraphic, printed, and customized artistic fonts. To ensure the quality of the training data, we implement a multi-step filtering pipeline where each raw image is subjected to the following rules:

\begin{itemize}[leftmargin=2em]
\item The image resolution must be at least $720 \times 720$.
\item The image must contain at least one text instance.
\item The MUSIQ~\cite{ke2021musiq} score must be at least 48 and the NRQM~\cite{ma2017learning} score must be at least 4 to ensure visual clarity.
\item The Q-Align~\cite{wu2023q} score must be at least 3 to ensure aesthetic quality.
\end{itemize}

After applying these strict filtering rules, the resulting images undergo professional OCR annotation. Specifically, ten dedicated domain experts annotate the filtered images, yielding a final collection of 39,986 high-quality images with precise OCR labels to form the SkyReels-OCR dataset. We then partition this collection by randomly sampling 200 images to constitute the SkyReels-OCR benchmark for evaluation, while the remaining images form the training set, denoted as SkyReels-OCR-40K. Note that a dataset size of $\sim 40\text{K}$ is empirically sufficient for effective LoRA fine-tuning of Qwen2.5-VL 7B.

\begin{figure}[t]
    \centering
    \includegraphics[width=\linewidth]{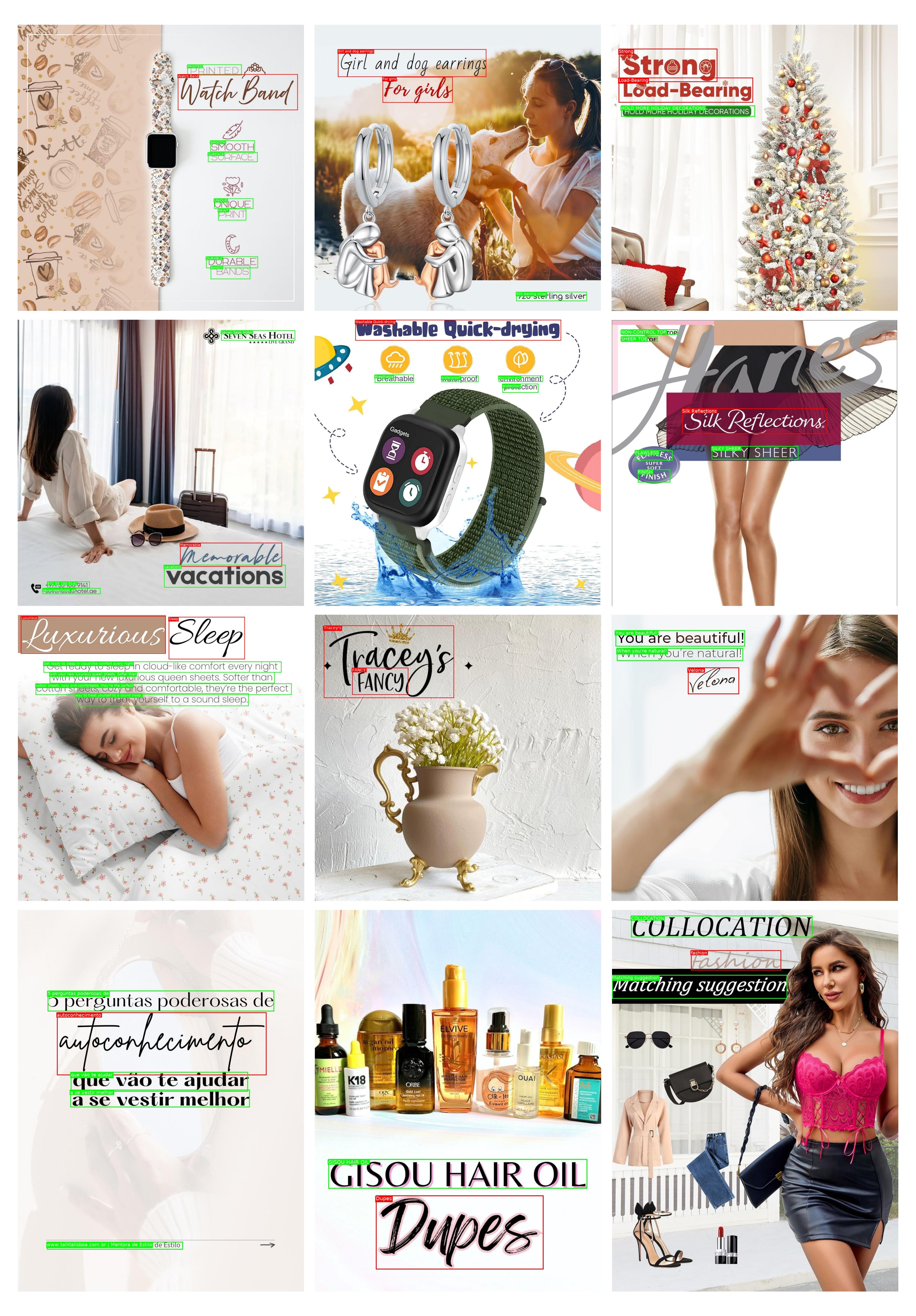}
    \caption{\textbf{Sample images from the SkyReels-OCR-40K dataset.} Green boxes denote regular fonts, while red boxes indicate non-regular fonts, including handwritten, calligraphic, and custom-designed artistic typefaces.}
    \label{fig:supplementary_1}
\end{figure}

\begin{table*}[h]
\centering
\caption{Statistics of text lines and characters for SkyReels-OCR-40K and AnyWord-3M.}
\label{tab:line_stats} % 推荐添加标签以便在正文中引用
\resizebox{\linewidth}{!}{
    \begin{tabular}{l|c|c|c|c|c|c|c}
    \toprule
    Dataset & image count & image w/o text & mean lines/img & mean chars/line & \#img $\le =5$ lines & \#line $\le =20$ chars & \#non-regular lines \\
    \midrule
    AnyWord-3M~\cite{tuo2024anytext} & 3.03M & 21.7K & 3.03 & 5.35 & 2.64M, 86.9\% & 9.15M, 99.6\% & - \\
    SkyReels-OCR-40K & 39.8K & 0 & 5.57 & 13.06 & 23.6K, 59.3\% & 0.18M, 81.8\% & 64.0K, 29.1\% \\
    \bottomrule
    \end{tabular}
}
\end{table*}

\noindent \textbf{SkyReels-OCR-40K.} Detailed statistics of text lines and characters for SkyReels-OCR-40K are presented in Table~\ref{tab:line_stats}, with AnyWord-3M~\cite{tuo2024anytext} included for comparison. Additionally, examples of annotated images from the dataset are provided in Figure~\ref{fig:supplementary_1}.

\begin{table*}[t]
\centering
\caption{Performance comparison of different OCR methods for regular and non-regular fonts on the SkyReels-OCR benchmark. \textbf{Boldface} and \underline{underlining} denote the best and second-best results, respectively.}
\label{tab:ocr_performance}
\begin{tabular}{l|ccc|ccc}
\toprule
\multirow{2}{*}{{Methods}} & \multicolumn{3}{c|}{{Regular Fonts}} & \multicolumn{3}{c}{{Non-Regular Fonts}} \\
 & Sen.~Acc$\uparrow$ & NED$\uparrow$ & Spatial$\uparrow$ & Sen.~Acc$\uparrow$ & NED$\uparrow$ & Spatial$\uparrow$ \\
\midrule
PP-OCRv4~\cite{ppocrv4} & 0.8769 & 0.9524 & 0.6543  & 0.3358 & 0.6914 & 0.4333 \\
PP-OCRv5~\cite{cui2025paddleocr30technicalreport} & 0.8635 & 0.9413 & 0.6548 & 0.5569 & 0.8200 & 0.5845 \\
PaddleOCR-VL~\cite{cui2025paddleocrvlboostingmultilingualdocument} & 0.8780 & 0.9110 & 0.4393 & 0.5444 & 0.6986 & 0.3920 \\
Qwen3-VL-8B-Instruct~\cite{qwen3vlgithub} & \textbf{0.9659} & \textbf{0.9906} & \underline{0.6887} & \underline{0.8925} & \underline{0.9709} & \underline{0.7831} \\
\midrule
Baseline & 0.8386 & 0.8921 & 0.4882 & 0.6319 & 0.8453 & 0.3665 \\
Ours  & \underline{0.9446} & \underline{0.9780} & \textbf{0.7617} & \textbf{0.9276} & \textbf{0.9824} & \textbf{0.8259} \\
\bottomrule
\end{tabular}
\end{table*}

\noindent \textbf{SkyReels-OCR Benchmark.} The SkyReels-OCR benchmark is utilized to evaluate OCR detection and recognition capabilities for regular and non-regular fonts using three metrics: Sen.~Acc, NED, and spatial IoU. For comparative analysis, we benchmark our model against several competitive methods: practical models PP-OCRv4~\cite{ppocrv4} and PP-OCRv5~\cite{cui2025paddleocr30technicalreport}, along with powerful VLM-based approaches including PaddleOCR-VL~\cite{cui2025paddleocrvlboostingmultilingualdocument}, Qwen2.5-VL-7B-Instruct~\cite{bai2025qwen25vltechnicalreport} (Baseline), and Qwen3-VL-8B-Instruct~\cite{qwen3vlgithub}. Quantitative comparisons are summarized in Table~\ref{tab:ocr_performance}. Qualitative results are shown in Figure~\ref{fig:supplementary_2}, where we present visual comparisons of different OCR models.

\begin{figure*}[t]
    \centering
    \includegraphics[width=\linewidth]{sec/figures/skyreels-poster_supplementary_2.jpg}
    \caption{\textbf{Sample images of regular and non-regular fonts from the SkyReels-OCR benchmark.} Best viewed at 300\% zoom.}
    \label{fig:supplementary_2}
\end{figure*}

\section{Content-Style Decoupled Dataset}
\label{sec:content_style_decoupled_dataset}
\noindent \textbf{Pipeline Details.} To collect high-quality data with paired target images and reference fonts, we design an automatic data generation pipeline consisting of three steps:

\textbf{(1) Divergent Text Generation:} We employ the Qwen3-8B~\cite{yang2025qwen3} LLM to generate semantically distinct replacement texts that adhere to the original sequence length and preserve the casing of each word, simulating a constrained text substitution task.

\textbf{(2) Reference Font Generation:} We utilize Nano Banana~\cite{gemini} to perform local text editing with the previously generated divergent texts, aiming for automated reference font generation. This approach maximally preserves the font style and color of the original text. Furthermore, to enable our model to concentrate on font style learning while ignoring background interference, we employ SAM2~\cite{ravi2024sam} to segment the text regions and then blend 
them with text-free background areas of other images.

\textbf{(3) Two-pronged Verification Strategy:} The final step in our content-style decoupling pipeline involves validating the fidelity of the local text edits. First, we employ our OCR model to confirm the content fidelity of the edits produced by Nano Banana. Specifically, we set a NED threshold of 0.9 between the ground truth replacement text and the edited text to retain only the successful instances. Subsequently, we leverage the DINO v2~\cite{oquab2023dinov2} model to verify stylistic consistency. This is achieved by setting a DINO similarity threshold of 0.8 to confirm that the edited text maintains the original style, ensuring successful style preservation during the editing process.

\begin{figure}[t]
    \centering
    \includegraphics[width=\linewidth]{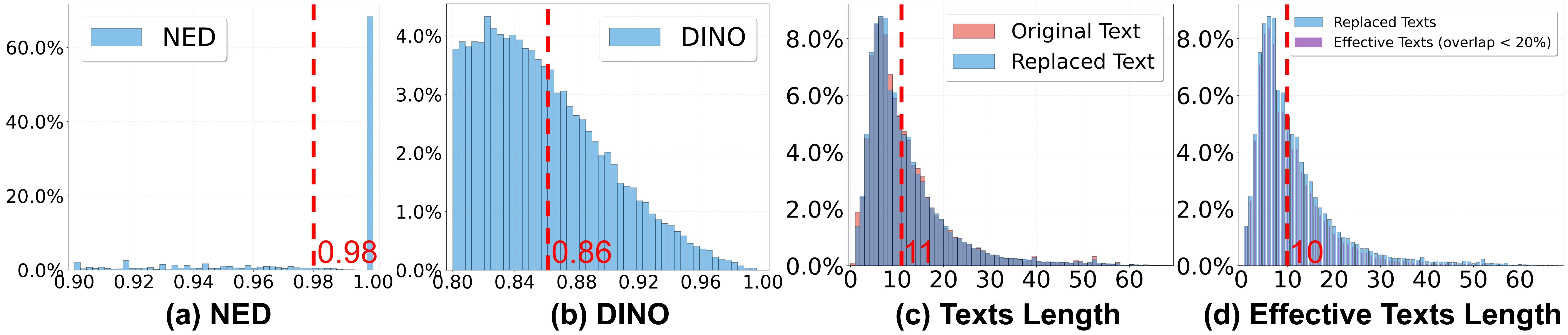}
    \caption{\textbf{Distributions of NED, DINO, text length, and effective text length.} The means are marked with red dashed lines.}
    \label{fig:supplementary_3}
\end{figure}

\noindent \textbf{Quantitative Analysis.} To comprehensively evaluate the quality of the constructed dataset, we report key quantitative statistics in Figure~\ref{fig:supplementary_3}. First, we verify the content accuracy and style consistency of the generated pairs by analyzing the NED and DINO distributions. Second, we confirm that the synthetic text lengths closely align with real-world distributions. Furthermore, to quantify the degree of decoupling, we define ``Effective Text'' as replacements with less than 20\% character overlap with the original text. Since the reference fonts merely provide style cues rather than semantic context, strict semantic alignment with the real posters is unnecessary; ensuring content divergence is sufficient. Our statistical analysis validates a high proportion of ``Effective Text'' in the dataset, demonstrating the success of our decoupling strategy.

\begin{figure}[t]
    \centering
    \includegraphics[width=\linewidth]{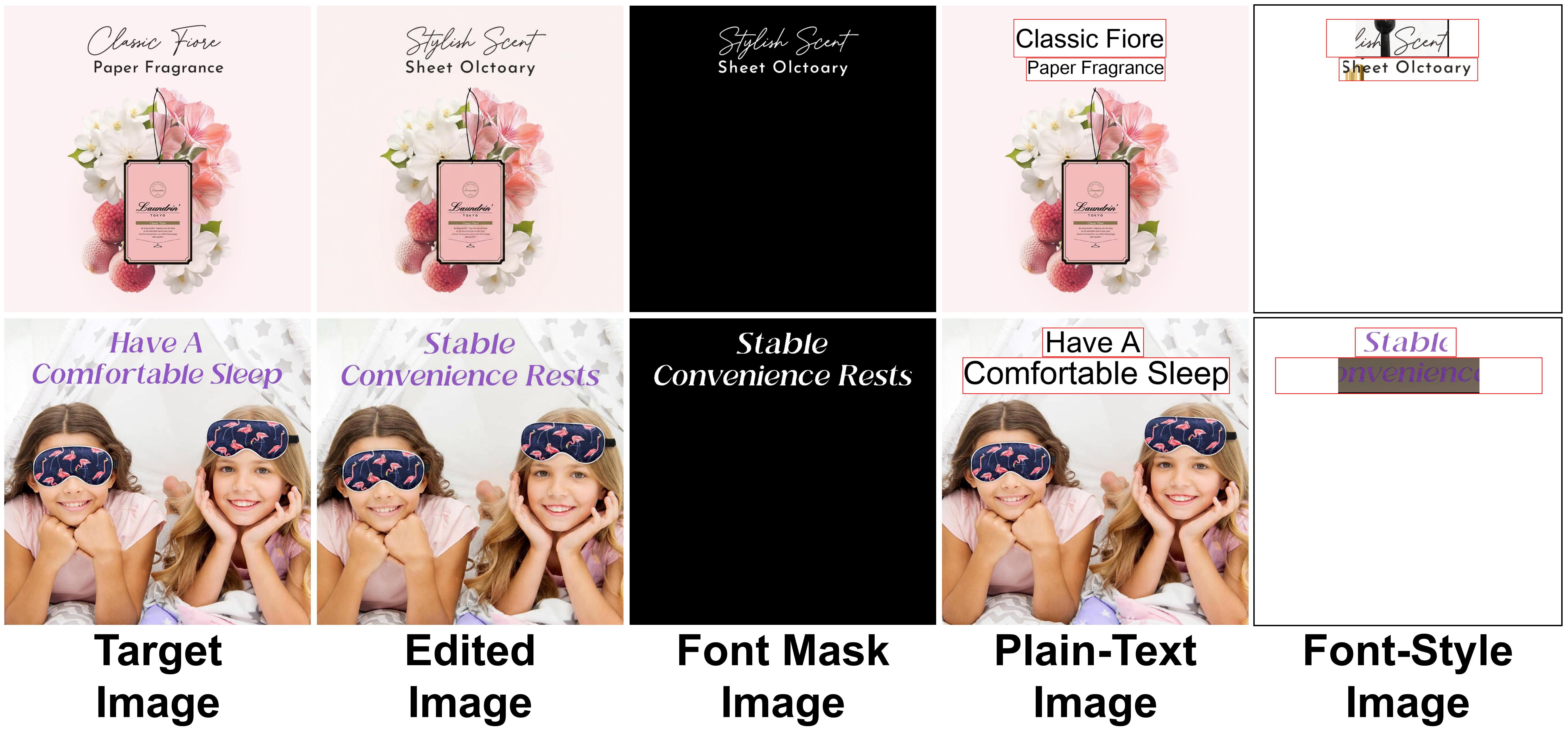}
    \caption{\textbf{Examples from the content-style decoupled dataset.} From left to right, the columns display: original target image, edited image with semantically divergent text, extracted font mask, plain-text layout, and font-style reference image. These examples clearly demonstrate precise content replacement while preserving typographic style.}
    \label{fig:supplementary_4}
\end{figure}

\noindent \textbf{Qualitative Check.} To qualitatively assess the dataset, we present visual samples of our training pairs in Figure~\ref{fig:supplementary_4}. These examples clearly demonstrate the high visual quality, precise content replacement, and accurate font style retention of our generated pairs. Note that the incomplete text in the style reference images is due to random cropping during data augmentation, which encourages the model to learn robust style patterns across varying text lengths.

\section{More Qualitative Results}
\label{sec:more_qualitative_results}

\begin{figure}[t]
    \centering
    \includegraphics[width=\linewidth]{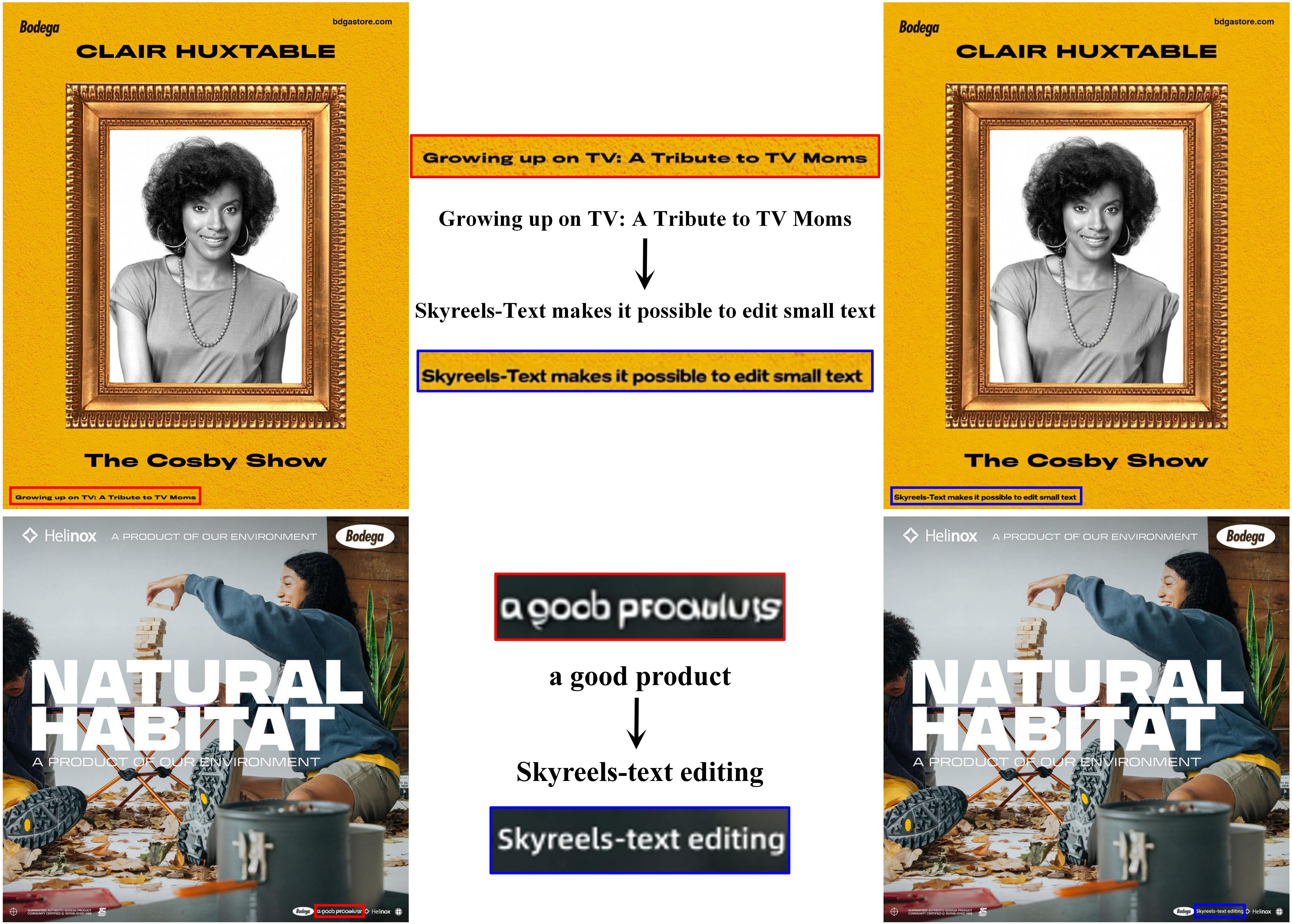}
    \caption{\textbf{Small-size text editing.} SkyReels-Text effectively replaces original small text or garbled artifacts (red boxes) with visually coherent target text (blue boxes), with zoomed-in details shown in the center.}
    \label{fig:supplementary_7}
\end{figure}

\noindent \textbf{Small-Size Text Editing.} In real-world workflows, users frequently need to edit small-size text regions in AI-generated posters, which often suffer from garbled or nonsensical characters. SkyReels-Text proves effective in tackling these highly challenging scenarios. As illustrated in Figure~\ref{fig:supplementary_7}, our model can generate or correct small text down to $\sim$20 pixels. Benefiting from the diverse text scales covered in our meticulously curated dataset, SkyReels-Text seamlessly replaces the original content---whether it is standard small text or garbled AI artifacts---with visually coherent target text. Furthermore, it effectively maintains the typographic identity of these small-size text regions and naturally integrates them into the surrounding background. This fine-grained control capability enhances the practical utility of our model.

\begin{figure}[t]
    \centering
    \includegraphics[width=\linewidth]{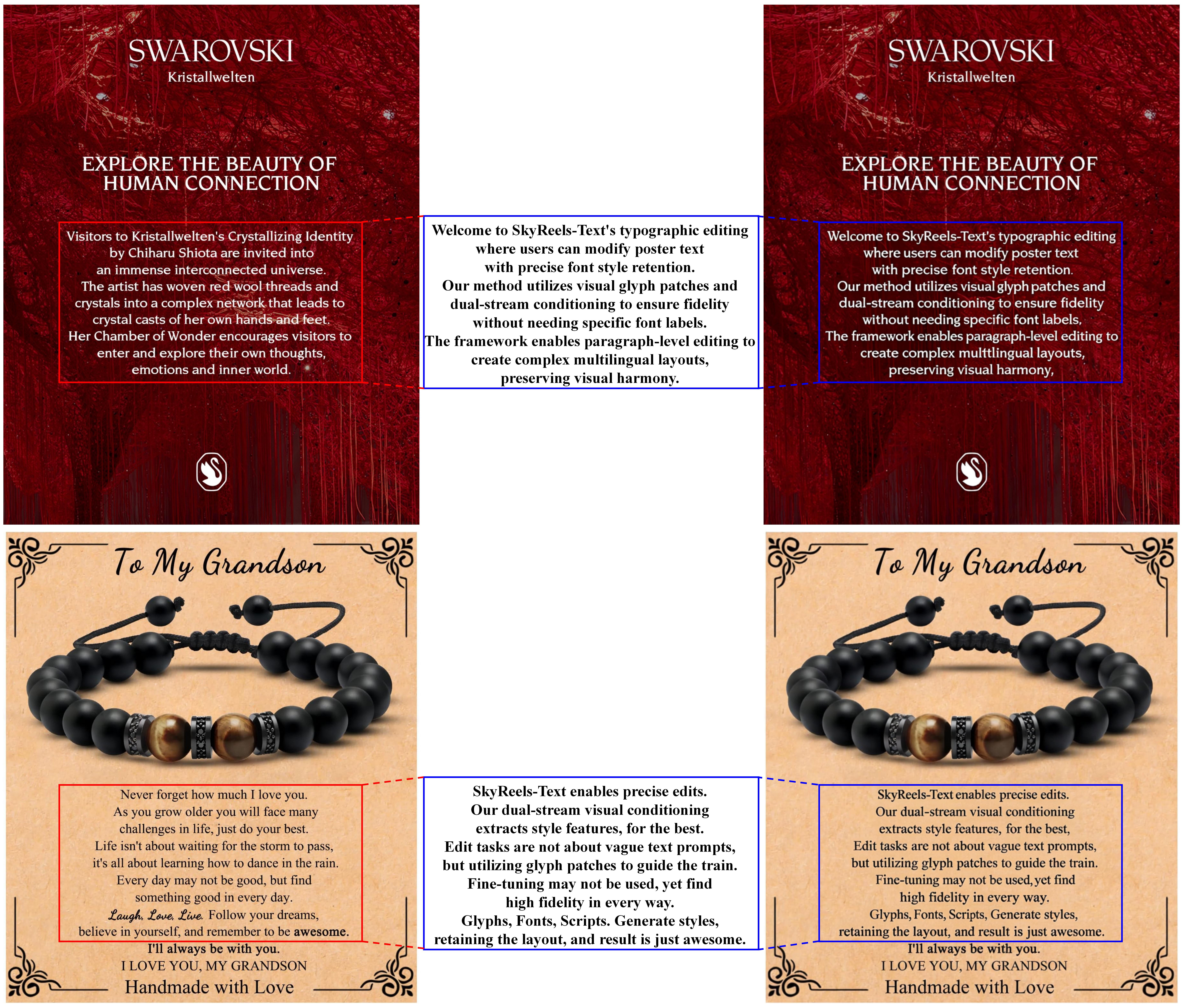}
    \caption{\textbf{Paragraph-level text editing.} The red and blue boxes indicate the original text and our edited results, respectively.}
    \label{fig:supplementary_8}
\end{figure}

\noindent \textbf{Paragraph-Level Text Editing.} Scaling up from word- and sentence-level edits, comprehensive poster redesign often necessitates the replacement of long paragraphs. To extend the capability of SkyReels-Text to this scale, we conducted continual training via LoRA on 10K paragraph-level samples (5K from our content-style decoupled dataset and 5K synthetic samples). As shown in Figure~\ref{fig:supplementary_8}, our model successfully handles paragraph-level editing for standard fonts. It effectively replaces the original multi-line text (red boxes) with substantial target content (blue boxes) while preserving the original style and layout. Although the model performs well on standard fonts, generating highly unconventional or artistic styles at the paragraph level remains challenging due to the scarcity of such dense, stylized data. We leave the exploration of this limitation to future work.

\begin{figure*}[t]
    \centering
    \includegraphics[width=0.9\linewidth]{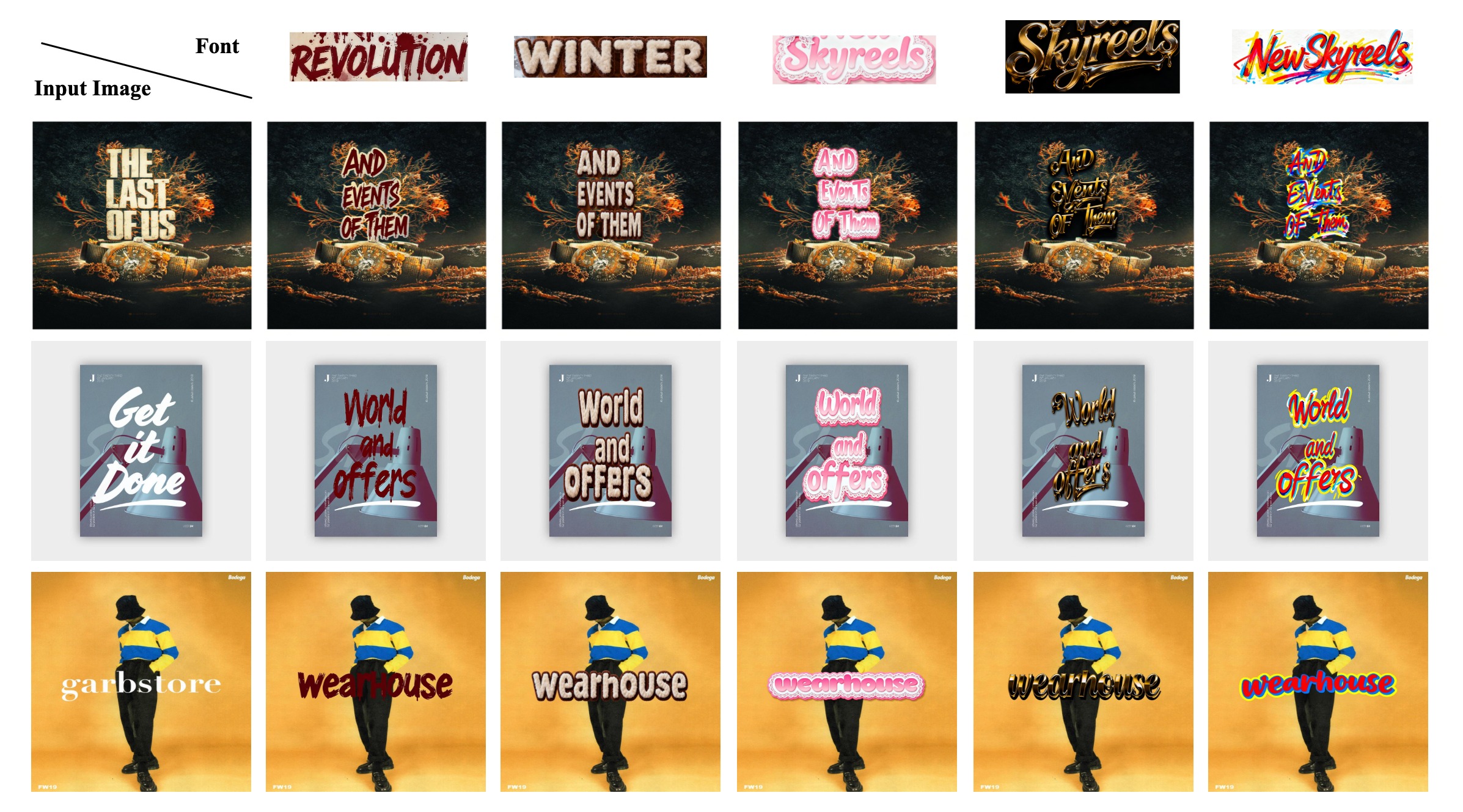}
    \caption{\textbf{More qualitative results of our model with single font styles.} The leftmost column shows the input images, and the top row displays the user-provided reference fonts.}
    \label{fig:single_comp_more_case}
\end{figure*}

\noindent \textbf{Text Editing with Single Font Styles.} Figure~\ref{fig:single_comp_more_case} presents more qualitative results of our model guided by a single reference font. Given an input image (leftmost column) and a user-provided glyph patch specifying the desired style (top row), our model accurately synthesizes the new textual content in the target regions. As demonstrated, SkyReels-Text successfully transfers complex stylistic nuances---such as rich material texture, 3D metallic gloss, and multi-chromatic brushwork---to the newly generated text. Furthermore, the model seamlessly blends the stylized text into the original layout while preserving the structural integrity and visual harmony of the unedited background. This highlights its powerful zero-shot font transfer capability, requiring no additional fine-tuning, which meets the rigorous demands of professional artistic design.

\begin{figure*}[t]
    \centering
    \includegraphics[width=0.9\linewidth]{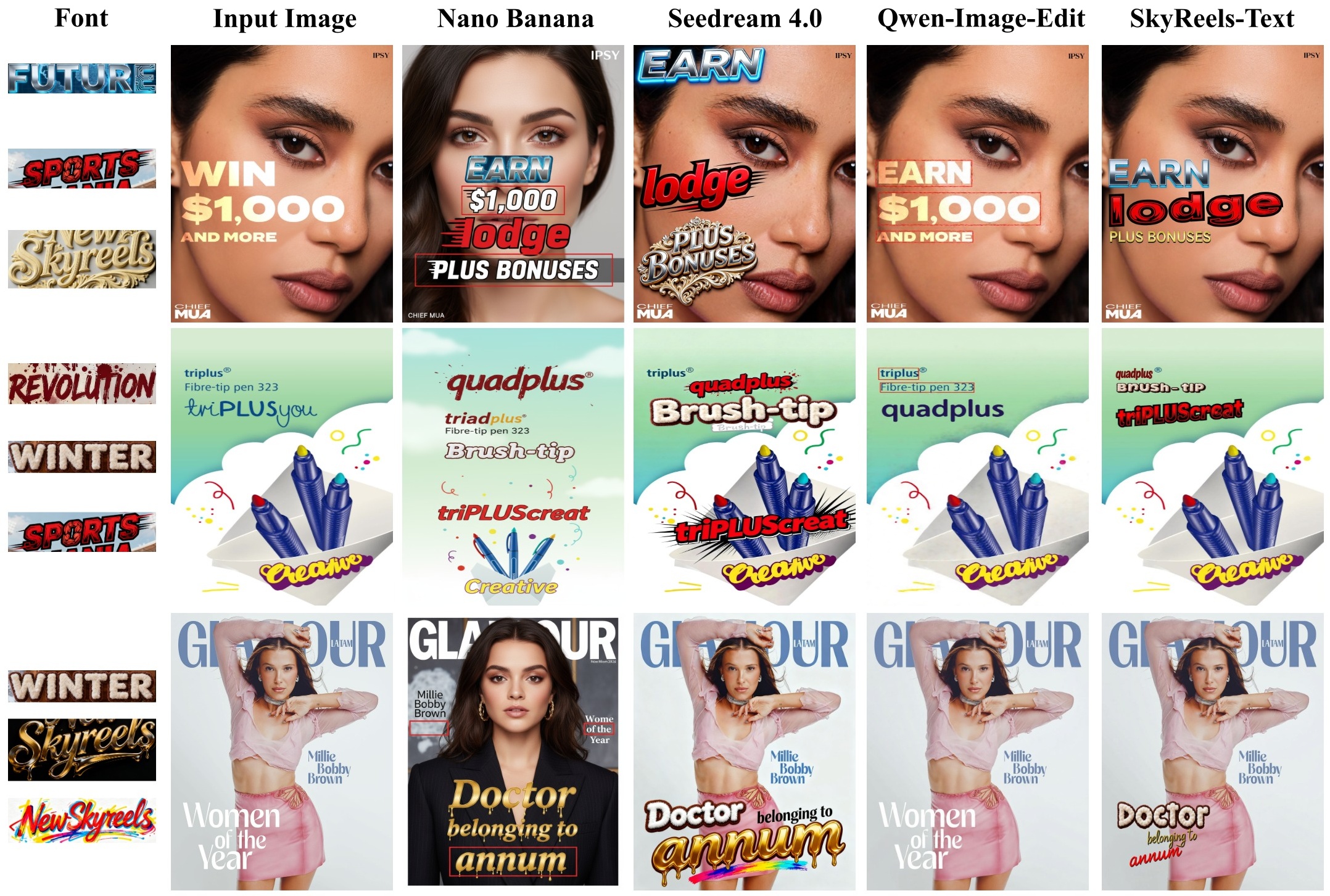}
    \caption{\textbf{Comparison with SOTA commercial image editing models in multi-font editing.} The first and second columns display the reference font styles and the input images, respectively. SkyReels-Text accurately applies distinct fonts to specific areas without mutual interference. Note that Flux Pro is unable to process multiple reference images; thus, it is omitted for a fair comparison.}
    \label{fig:multicomp}
\end{figure*}

\noindent \textbf{Text Editing with Multiple Font Styles.} Most existing image or text editing methods require multiple rounds of interaction to apply different font styles across text regions. In contrast, SkyReels-Text can simultaneously edit multiple regions with distinct font styles in a single inference round, achieving precise and efficient multi-style editing. Figure~\ref{fig:multicomp} illustrates the capability of our model to edit multiple text regions with diverse font styles, in comparison to SOTA image editing models. The results show that Nano Banana produces severe artifacts and distorts the original image structure, Qwen-Image-Edit tends to leave the image unedited, and Seedream~4.0 fails to preserve either the correct text positioning or the font style. Conversely, SkyReels-Text achieves accurate, style-consistent, and spatially faithful edits in a single inference pass, which takes $\sim$8 seconds on a single A800 GPU.

\noindent \textbf{Handwritten Text Generation.} Tables~\ref{tab:English_handwrite} and \ref{tab:Chinese_handwrite} illustrate the capability of SkyReels-Text to generate English and Chinese handwritten text. Existing methods for handwritten text generation are typically monolingual---requiring separate training for English and Chinese, with one model per language. In contrast, SkyReels-Text enables zero-shot multilingual generation: given only a style reference image, it synthesizes high-fidelity handwritten text without any additional training. This strong generalization capability eliminates the need for language-specific fine-tuning, offering a flexible, efficient, and scalable solution for style-controllable text image synthesis across diverse languages.

\begin{table*}[h]
  \centering
  \caption{English handwritten text generation.}
  \label{tab:English_handwrite}
  % 左侧标题列 + 单栏图片展示
  \begin{tabular}{l l}  
    \toprule
    Ground Truth & 
    \parbox{0.7\linewidth}{\raggedright  % 单栏宽度调整为0.7行宽
      \resizebox{\linewidth}{!}{\includegraphics{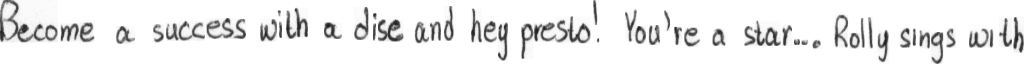}}\\[0.1cm]
    } \\
    Style sample & 
    \parbox{0.7\linewidth}{\raggedright  % 单栏宽度调整为0.7行宽
      \resizebox{\linewidth}{!}{\includegraphics{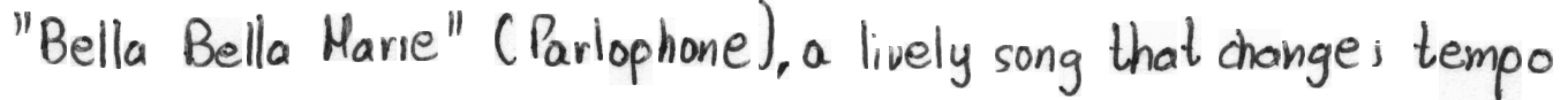}}\\[0.1cm]
    } \\
    \textbf{SkyReels-Text} & 
    \parbox{0.7\linewidth}{\raggedright
      \resizebox{\linewidth}{!}{\includegraphics{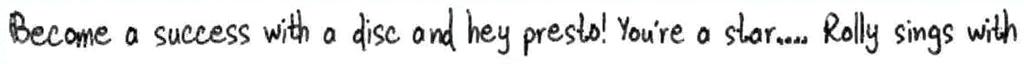}}\\[0.1cm]
    } \\
    \midrule
    Ground Truth & 
    \parbox{0.7\linewidth}{\raggedright  % 单栏宽度调整为0.7行宽
      \resizebox{\linewidth}{!}{\includegraphics{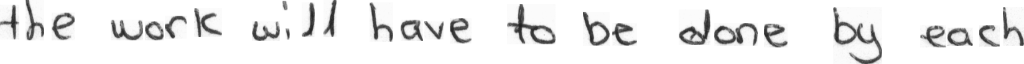}}\\[0.1cm]
    } \\
    Style sample & 
    \parbox{0.7\linewidth}{\raggedright  % 单栏宽度调整为0.7行宽
      \resizebox{\linewidth}{!}{\includegraphics{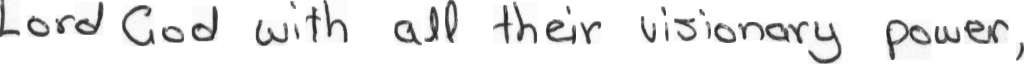}}\\[0.1cm]
    } \\
    \textbf{SkyReels-Text} & 
    \parbox{0.7\linewidth}{\raggedright
      \resizebox{\linewidth}{!}{\includegraphics{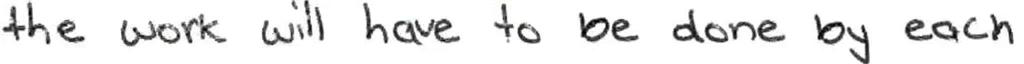}}\\[0.1cm]
    } \\
    \midrule
    % 标题行仅保留一栏文本
    Ground Truth & 
    % \parbox{0.7\linewidth}{\small 这一研究成果对准确推算地球生命的历史非常有帮助} \\
    \parbox{0.7\linewidth}{\raggedright  % 单栏宽度调整为0.7行宽
      \resizebox{\linewidth}{!}{\includegraphics{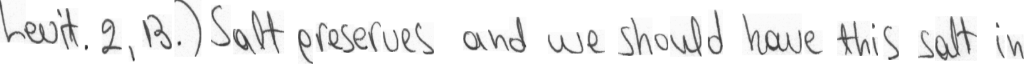}}\\[0.1cm]
    } \\
    Style sample & 
    \parbox{0.7\linewidth}{\raggedright  % 单栏宽度调整为0.7行宽
      \resizebox{\linewidth}{!}{\includegraphics{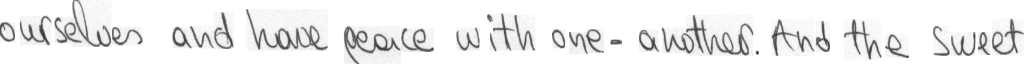}}\\[0.1cm]
    } \\
    \textbf{SkyReels-Text} & 
    \parbox{0.7\linewidth}{\raggedright
      \resizebox{\linewidth}{!}{\includegraphics{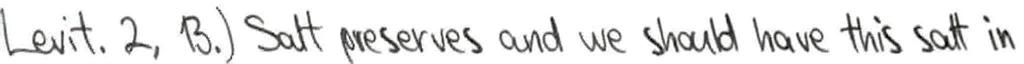}}\\[0.1cm]
    } \\
    \midrule
    Ground Truth & 
    \parbox{0.7\linewidth}{\raggedright  % 单栏宽度调整为0.7行宽
      \resizebox{\linewidth}{!}{\includegraphics{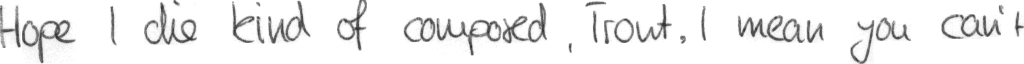}}\\[0.1cm]
    } \\
    Style sample & 
    \parbox{0.7\linewidth}{\raggedright  % 单栏宽度调整为0.7行宽
      \resizebox{\linewidth}{!}{\includegraphics{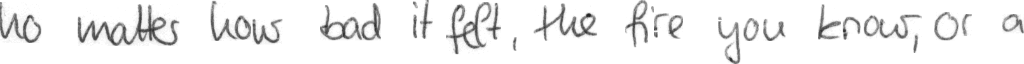}}\\[0.1cm]
    } \\
    \textbf{SkyReels-Text} & 
    \parbox{0.7\linewidth}{\raggedright
      \resizebox{\linewidth}{!}{\includegraphics{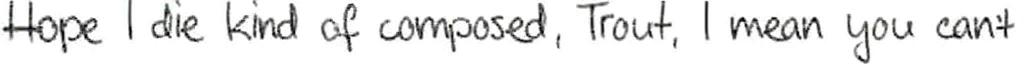}}\\[0.1cm]
    } \\
    \midrule
    Ground Truth & 
    \parbox{0.7\linewidth}{\raggedright  % 单栏宽度调整为0.7行宽
      \resizebox{\linewidth}{!}{\includegraphics{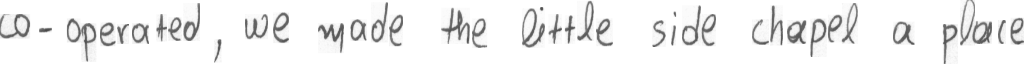}}\\[0.1cm]
    } \\
    Style sample & 
    \parbox{0.7\linewidth}{\raggedright  % 单栏宽度调整为0.7行宽
      \resizebox{\linewidth}{!}{\includegraphics{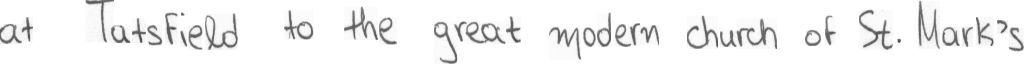}}\\[0.1cm]
    } \\
    \textbf{SkyReels-Text} & 
    \parbox{0.7\linewidth}{\raggedright
      \resizebox{\linewidth}{!}{\includegraphics{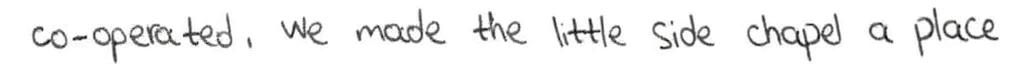}}\\[0.1cm]
    } \\
    \midrule
    Ground Truth & 
    \parbox{0.7\linewidth}{\raggedright  % 单栏宽度调整为0.7行宽
      \resizebox{\linewidth}{!}{\includegraphics{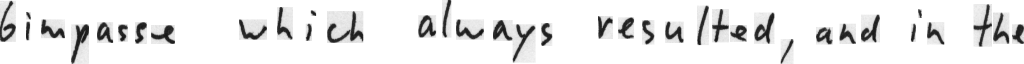}}\\[0.1cm]
    } \\
    Style sample & 
    \parbox{0.7\linewidth}{\raggedright  % 单栏宽度调整为0.7行宽
      \resizebox{\linewidth}{!}{\includegraphics{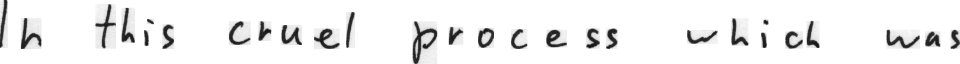}}\\[0.1cm]
    } \\
    \textbf{SkyReels-Text} & 
    \parbox{0.7\linewidth}{\raggedright
      \resizebox{\linewidth}{!}{\includegraphics{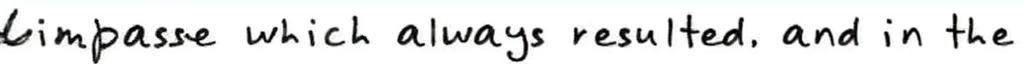}}\\[0.1cm]
    } \\
    \midrule
    Ground Truth & 
    \parbox{0.7\linewidth}{\raggedright  % 单栏宽度调整为0.7行宽
      \resizebox{\linewidth}{!}{\includegraphics{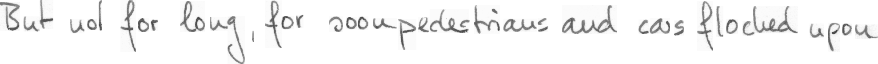}}\\[0.1cm]
    } \\
    Style sample & 
    \parbox{0.7\linewidth}{\raggedright  % 单栏宽度调整为0.7行宽
      \resizebox{\linewidth}{!}{\includegraphics{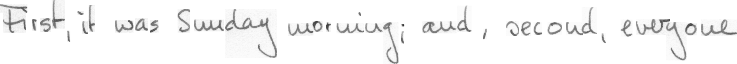}}\\[0.1cm]
    } \\
    \textbf{SkyReels-Text} & 
    \parbox{0.7\linewidth}{\raggedright
      \resizebox{\linewidth}{!}{\includegraphics{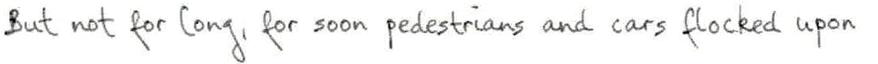}}\\[0.1cm]
    } \\
    \bottomrule
  \end{tabular}
\end{table*}

\begin{table*}[h]
  \centering
  \caption{Chinese handwritten text generation.}
  \label{tab:Chinese_handwrite}
  % 左侧标题列 + 单栏图片展示
  \begin{tabular}{l l}  
    \toprule
    % 标题行仅保留一栏文本
    Ground Truth & 
    % \parbox{0.7\linewidth}{\small 这一研究成果对准确推算地球生命的历史非常有帮助} \\
    \parbox{0.7\linewidth}{\raggedright  % 单栏宽度调整为0.7行宽
      \resizebox{\linewidth}{!}{\includegraphics{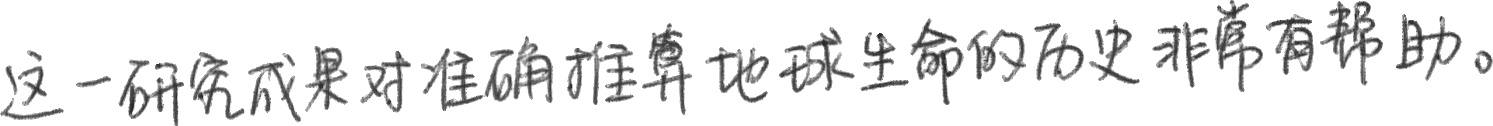}}\\[0.1cm]
    } \\
    Style sample & 
    \parbox{0.7\linewidth}{\raggedright  % 单栏宽度调整为0.7行宽
      \resizebox{\linewidth}{!}{\includegraphics{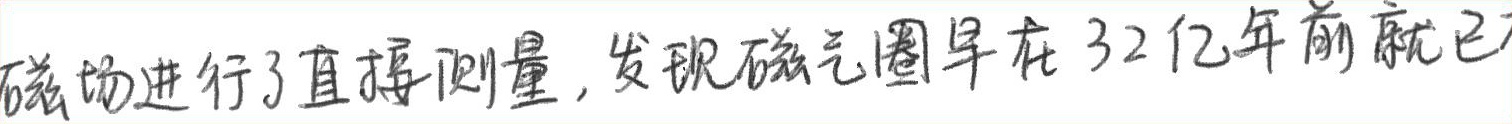}}\\[0.1cm]
    } \\
    \textbf{SkyReels-Text} & 
    \parbox{0.7\linewidth}{\raggedright
      \resizebox{\linewidth}{!}{\includegraphics{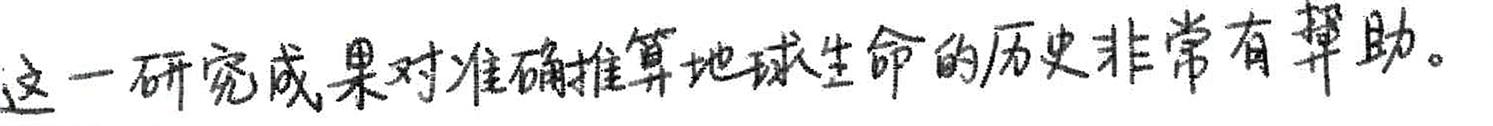}}\\[0.1cm]
    } \\
    \midrule
    Ground Truth & 
    \parbox{0.7\linewidth}{\raggedright  % 单栏宽度调整为0.7行宽
      \resizebox{\linewidth}{!}{\includegraphics{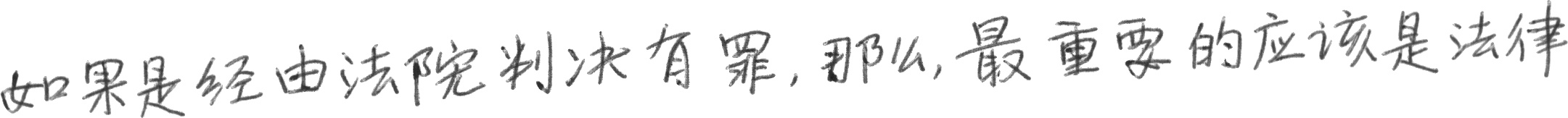}}\\[0.1cm]
    } \\
    Style sample & 
    \parbox{0.7\linewidth}{\raggedright  % 单栏宽度调整为0.7行宽
      \resizebox{\linewidth}{!}{\includegraphics{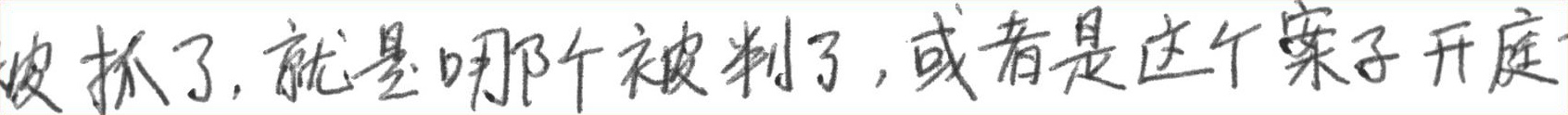}}\\[0.1cm]
    } \\
    \textbf{SkyReels-Text} & 
    \parbox{0.7\linewidth}{\raggedright
      \resizebox{\linewidth}{!}{\includegraphics{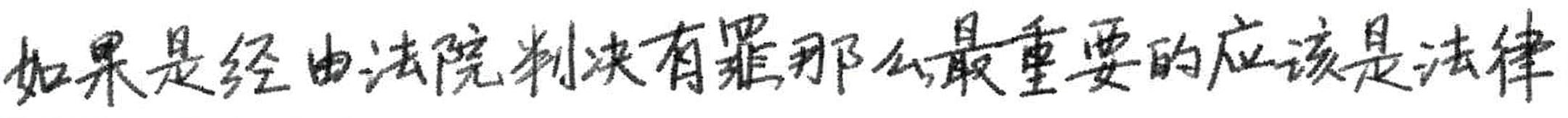}}\\[0.1cm]
    } \\
    \midrule
    Ground Truth & 
    \parbox{0.7\linewidth}{\raggedright  % 单栏宽度调整为0.7行宽
      \resizebox{\linewidth}{!}{\includegraphics{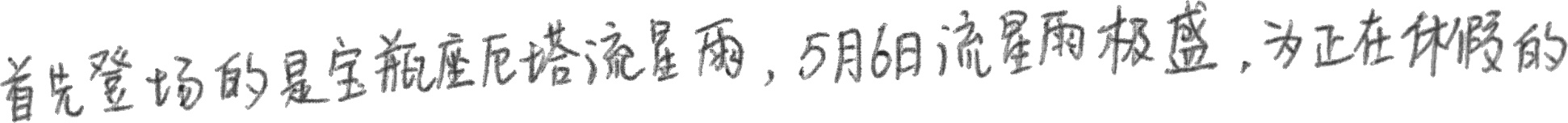}}\\[0.1cm]
    } \\
    Style sample & 
    \parbox{0.7\linewidth}{\raggedright  % 单栏宽度调整为0.7行宽
      \resizebox{\linewidth}{!}{\includegraphics{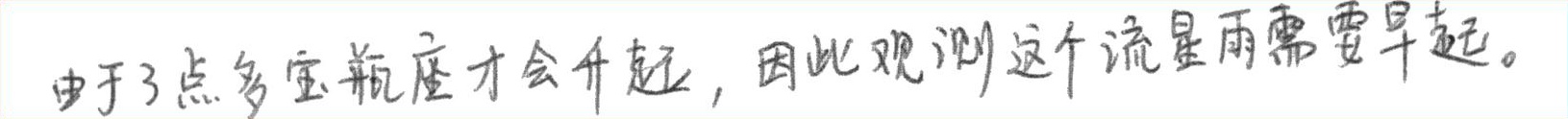}}\\[0.1cm]
    } \\
    \textbf{SkyReels-Text} & 
    \parbox{0.7\linewidth}{\raggedright
      \resizebox{\linewidth}{!}{\includegraphics{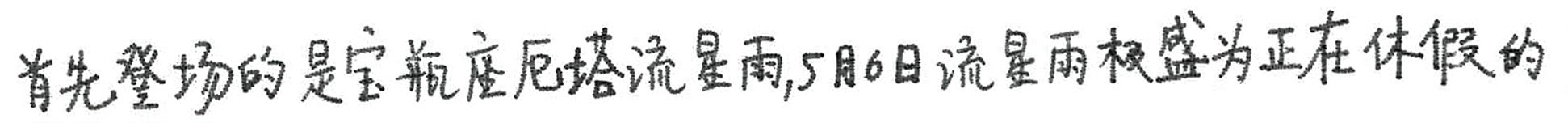}}\\[0.1cm]
    } \\
    \midrule
    Ground Truth & 
    \parbox{0.7\linewidth}{\raggedright  % 单栏宽度调整为0.7行宽
      \resizebox{\linewidth}{!}{\includegraphics{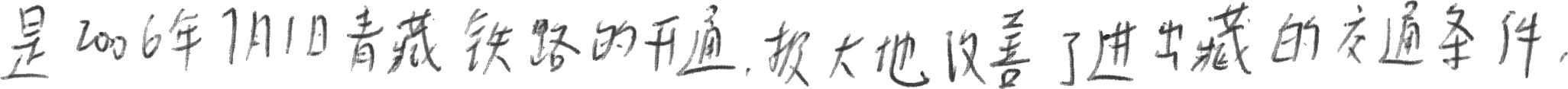}}\\[0.1cm]
    } \\
    Style sample & 
    \parbox{0.7\linewidth}{\raggedright  % 单栏宽度调整为0.7行宽
      \resizebox{\linewidth}{!}{\includegraphics{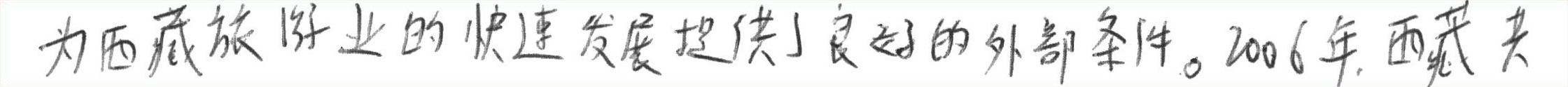}}\\[0.1cm]
    } \\
    \textbf{SkyReels-Text} & 
    \parbox{0.7\linewidth}{\raggedright
      \resizebox{\linewidth}{!}{\includegraphics{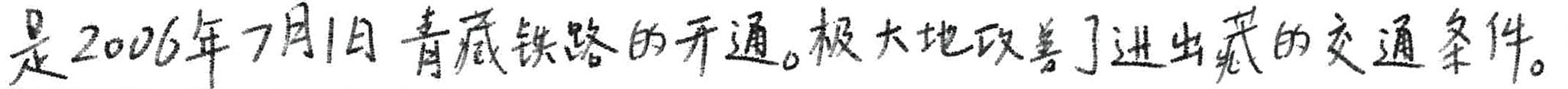}}\\[0.1cm]
    } \\
    \midrule
    Ground Truth & 
    \parbox{0.7\linewidth}{\raggedright  % 单栏宽度调整为0.7行宽
      \resizebox{\linewidth}{!}{\includegraphics{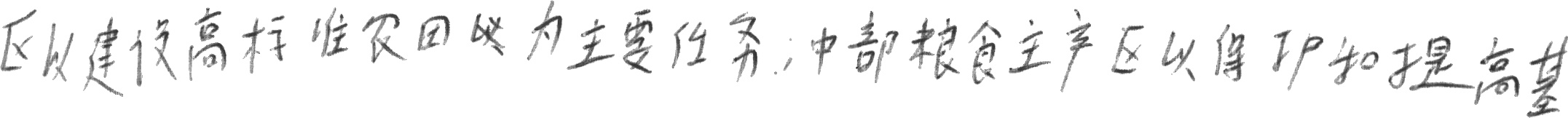}}\\[0.1cm]
    } \\
    Style sample & 
    \parbox{0.7\linewidth}{\raggedright  % 单栏宽度调整为0.7行宽
      \resizebox{\linewidth}{!}{\includegraphics{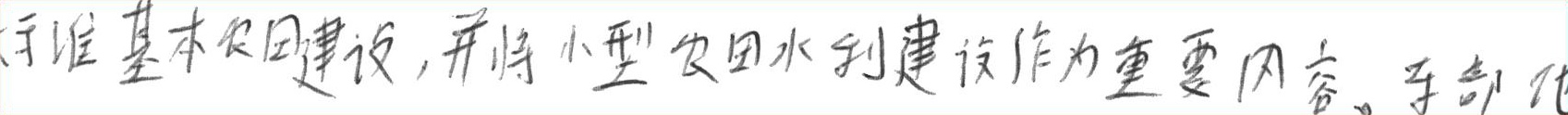}}\\[0.1cm]
    } \\
    \textbf{SkyReels-Text} & 
    \parbox{0.7\linewidth}{\raggedright
      \resizebox{\linewidth}{!}{\includegraphics{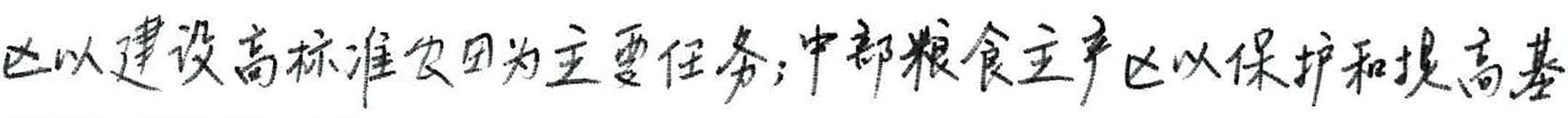}}\\[0.1cm]
    } \\
    \midrule
     Ground Truth & 
    \parbox{0.7\linewidth}{\raggedright  % 单栏宽度调整为0.7行宽
      \resizebox{\linewidth}{!}{\includegraphics{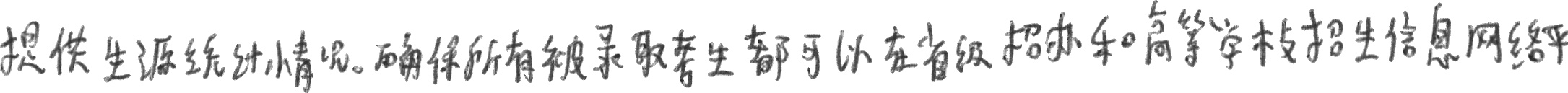}}\\[0.1cm]
    } \\
    Style sample & 
    \parbox{0.7\linewidth}{\raggedright  % 单栏宽度调整为0.7行宽
      \resizebox{\linewidth}{!}{\includegraphics{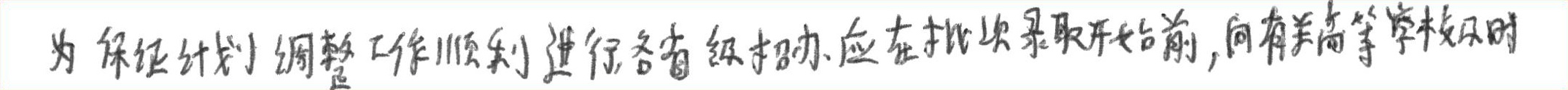}}\\[0.1cm]
    } \\
    \textbf{SkyReels-Text} & 
    \parbox{0.7\linewidth}{\raggedright
      \resizebox{\linewidth}{!}{\includegraphics{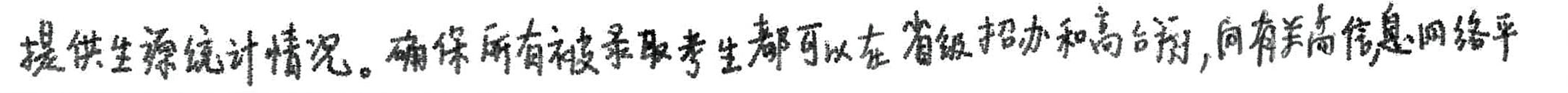}}\\[0.1cm]
    } \\
    \midrule
    Ground Truth & 
    \parbox{0.7\linewidth}{\raggedright  % 单栏宽度调整为0.7行宽
      \resizebox{\linewidth}{!}{\includegraphics{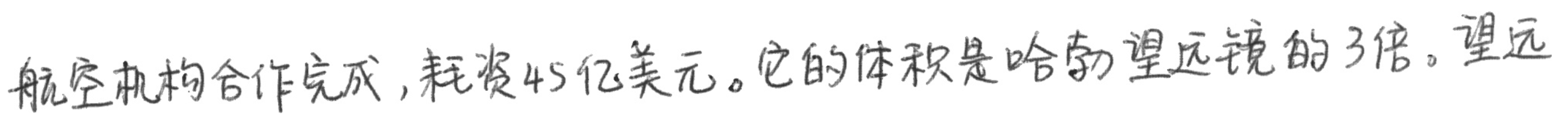}}\\[0.1cm]
    } \\
    Style sample & 
    \parbox{0.7\linewidth}{\raggedright  % 单栏宽度调整为0.7行宽
      \resizebox{\linewidth}{!}{\includegraphics{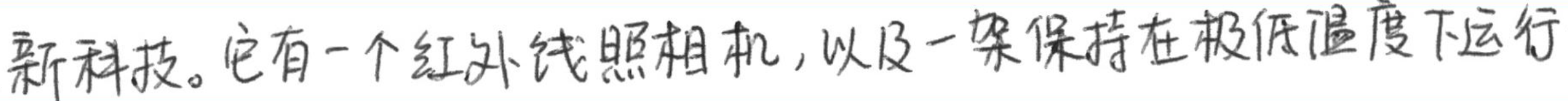}}\\[0.1cm]
    } \\
    \textbf{SkyReels-Text} & 
    \parbox{0.7\linewidth}{\raggedright
      \resizebox{\linewidth}{!}{\includegraphics{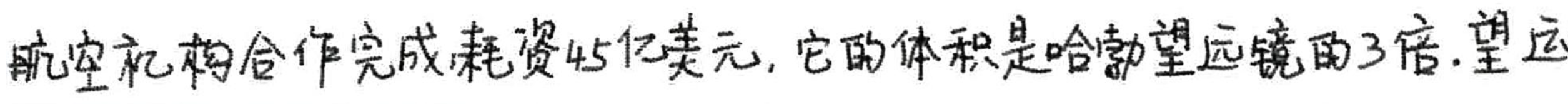}}\\[0.1cm]
    } \\
    \bottomrule
  \end{tabular}
\end{table*}

\end{document}